\documentclass{article} 
\usepackage{iclr2027_conference,times}

\usepackage{amsmath,amsfonts,bm}

\def\eqref#1{equation~\ref{#1}}

\def\1{\bm{1}}

\DeclareMathAlphabet{\mathsfit}{\encodingdefault}{\sfdefault}{m}{sl}
\SetMathAlphabet{\mathsfit}{bold}{\encodingdefault}{\sfdefault}{bx}{n}

\usepackage{url}

\usepackage{booktabs}
\usepackage{multirow}
\usepackage{amssymb}
\usepackage{graphicx} 
\usepackage{float}
\usepackage{array}
\usepackage{microtype}
\usepackage{placeins}
\usepackage{hyperref}
\usepackage{tablefootnote} 


\newcommand{\glsentry}[2]{%
    \par\noindent\textbf{#1.} #2\par\vspace{0.35em}
}

\title{Decodable but Misrouted: Sparse Features Uncover a Readout Gap in Vision-Language Models for Harmful Meme Detection}

\author{Girish A.~Koushik\thanks{Corresponding Author} , Diptesh Kanojia
\\
Nature-Inspired Computing \& Engineering \\
Department of Computer Science\\
University of Surrey\\
Guildford, United Kingdom \\
\texttt{\{g.koushik,d.kanojia\}@surrey.ac.uk} \\
\And
Helen Treharne \\
Surrey Centre for Cyber Security \\
Department of Computer Science \\
University of Surrey \\
Guildford, United Kingdom \\
\texttt{\{h.treharne\}@surrey.ac.uk} \\
}

\iclrfinalcopy 

\begin{document}
\addtocontents{toc}{\protect\setcounter{tocdepth}{-1}}

\maketitle

\thispagestyle{empty} 
\pagestyle{plain}

\raggedbottom

\begin{abstract}
When large vision-language models misclassify harmful memes, the failure may reflect missing internal evidence or an inability to route represented evidence to their outputs. We distinguish these cases in Gemma-3 and Qwen3.5 using sparse autoencoders, role-conditioned probes, causal interventions, and recovery experiments across six harmful content benchmarks, with additional Spanish and Hindi-English code-mixed evaluations. Sparse readouts outperform native prediction on all six primary binary tasks: Qwen averages $0.740$ versus $0.432$ for native macro-F1, residual reconstruction reaches $0.486$, and Gemma improves from $0.532$ to $0.714$. These gains measure how accessible the label is to a supervised readout; they do not show that the model's native generation already applies such a decision rule. The most informative token role also varies by task. Under the evaluated scales, Qwen silent-feature ablation is \(24\text{--}63\) times more probe-sensitive, whereas routed-feature patching on literal yes/no tasks is \(16\text{--}140\) times more output-sensitive. Native-only threshold calibration explains much, but not all of the gap: on five tasks with matched probe scores, it recovers $69.8$\% of the raw native-to-probe difference, while direct routing adds $0.094$ mean macro-F1 beyond calibrated native scoring. Joint gold-label, probe-KL, and pairwise LoRA supervision improves dedicated FHM prediction, but a gold-only adapter performs better on the shared seven-task mean. A case study of Gemma-3-12B on the Facebook Hateful Memes dataset finds a distributed rank-32 image-prompt interaction, reaching $0.756$ versus $0.685$ native macro-F1. Robustness controls show that the signal is not explained solely by accompanying OCR and depends on paired visual evidence, and that it extends beyond English. In many of the errors we study, the evidence is represented but does not reach the answer; therefore, routing rather than representation alone is a common bottleneck in harmful meme classification.

\textbf{Disclaimer}: This paper includes references to content that some readers may find triggering or upsetting, consistent with the nature of the task.
\end{abstract}

\section{Introduction}      \label{sec:intro}

Memes compress visual content, written language, social context, and cultural implication into a single communicative unit. Their harmfulness often comes from how the image and text combine, even when neither is harmful on its own. The same image can become hateful under a different caption, while the same caption can become benign or harmful under a different visual context. Benchmarks such as the Facebook Hateful Memes (FHM) Challenge make this ambiguity explicit through benign confounders designed to defeat unimodal priors~\citep{kiela2020hateful}. Yet a nominally multimodal model can score well without having learned the relevant cross-modal relationship~\citep{hessel-lee-2020-multimodal}. This makes harmful meme detection a hard test case: we can ask whether a model gets the label right, and also what information its internal computation represents and which of it the model actually uses.

Large vision-language models (LVLMs) are increasingly used as general-purpose classifiers through prompting, constrained decoding, or label scoring. When such a model returns an incorrect harmfulness label, however, the output alone leaves two distinct explanations unresolved. The model may have failed to represent the relevant evidence, or it may have represented that evidence internally but failed to route it into the final decision. The two cases call for different fixes. A representational failure requires better perception, multimodal integration, or training on more data. A readout failure instead suggests that the information is already available and may be recoverable through a better decision pathway. Related discrepancies between internal encoding and external behavior have been observed in language models~\citep{orgad2025llmsknowshowintrinsic}, and concurrent work reports them for clinical triage decisions and omnimodal premise-conflict detection~\citep{fraile2026internal,quang2026senses}, but they remain poorly understood for multimodal safety judgments.

Sparse autoencoders (SAEs) offer a useful coordinate system for separating these possibilities. By decomposing dense activations into sparse feature directions, SAEs enable training lightweight readouts, comparing feature importance with output alignment, and intervening on selected directions~\citep{bricken2023towards,ICLR2024_1fa1ab11,lieberum-etal-2024-gemma,marks2025sparse}. Predictive sparse features are not automatically causal: an interpretable probe does not show that the native model uses the same features. A complete account must connect four levels of evidence: whether harmful-content information is decodable, where it appears across token roles, whether the relevant directions influence the native output, and whether the signal can be converted into improved model behavior.

We study these questions across Gemma-3 and Qwen3.5 model families using six benchmarks covering hate speech, harmfulness, misogyny, and offensiveness, followed by Spanish and Hindi-English code-mixed robustness experiments. We compare native constrained prediction with residual-reconstruction hooks and supervised readouts over base, residual, and cross-layer sparse representations. We then localize the signal across prompt, image, generated, and full-sequence states, contrasting probe-important features with independently selected output-routed features and testing both sets through causal interventions. Finally, we evaluate whether the recovered signal can be rerouted directly at the output or distilled into the LVLM itself.



\paragraph{Contributions.}
(i) An evaluation framework that holds the frozen LVLM and examples fixed while comparing native, reconstruction-hook, sparse-probe, and output-routed readouts; with it, we find a readout gap on all six binary tasks in two model families, with task-dependent token roles (Section~\ref{sec:internal_repr}). (ii) Causal evidence that, in Qwen, the SAE directions that discriminate harmful content are largely separate from those that control the native answer, with a geometric replication in Gemma (Section~\ref{subsec:mechanism}). (iii) A decomposition of the gap into native threshold miscalibration and missing ranking signal, together with output routing and jointly supervised LoRA evaluated against gold-only controls (Section~\ref{sec:latent_signal}). (iv) Evidence for a distributed low-rank image--prompt interaction on FHM in Gemma-3-12B (Section~\ref{subsec:fhm_bilinear}), and robustness checks showing that the gap extends to Spanish and Hindi-English memes and depends on the paired image rather than supplied OCR alone.

\begin{figure}[t]
    \centering
    \includegraphics[width=0.95\linewidth]{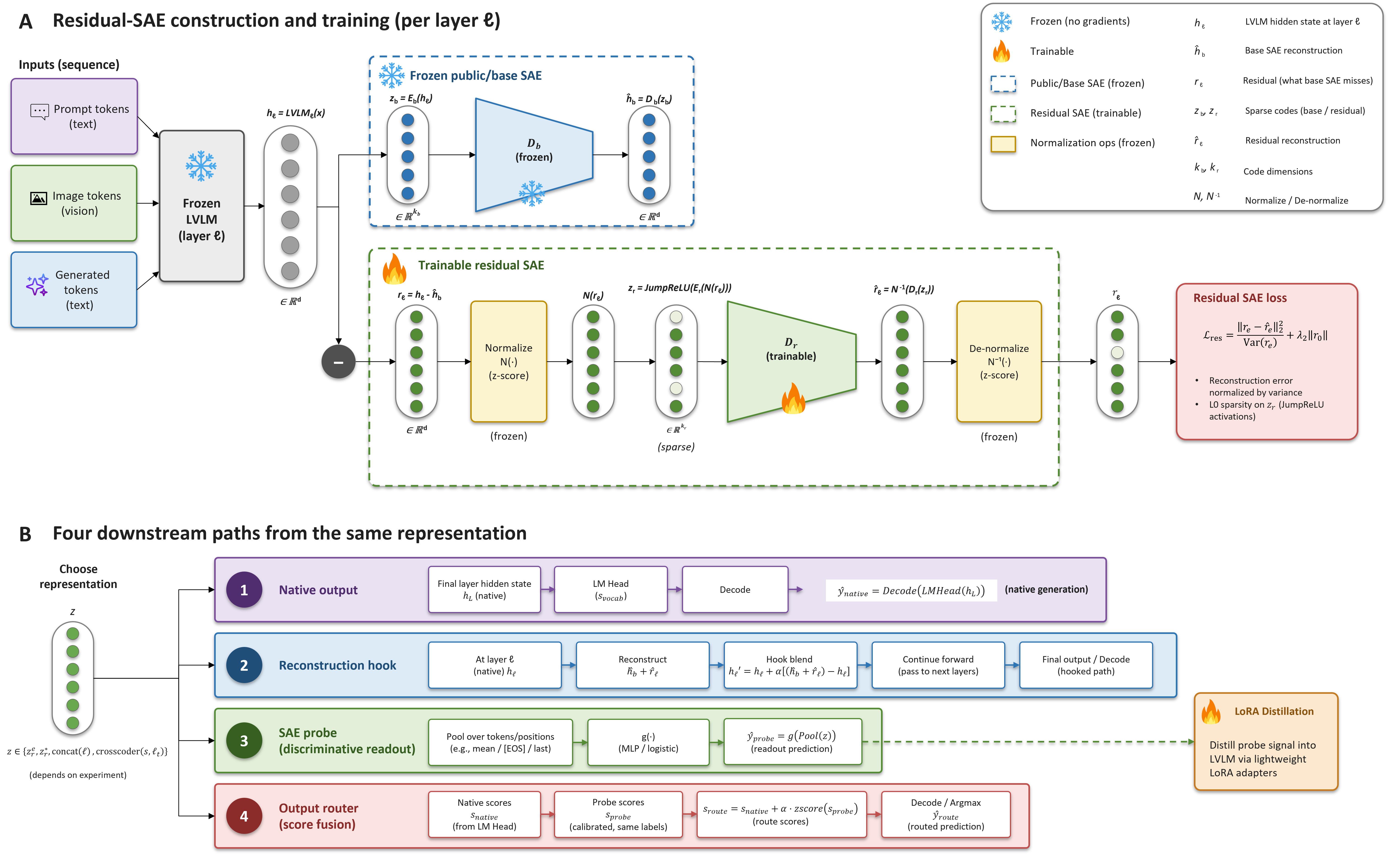}
    \caption{\textbf{Residual-SAE construction and readout pathways.} \textbf{A.} A frozen base SAE reconstructs layer-\(\ell\) states, while a trainable SAE models the normalized reconstruction residual, yielding base and joint reconstructions. \textbf{B.} The resulting representation supports native scoring, residual-reconstruction hooks, sparse probing, calibrated output routing, and jointly supervised LoRA. Snowflakes and flames denote frozen and trainable components, respectively.}
    \label{fig:residual_sae}
\end{figure}


\section{Background and Experimental Framework}     \label{sec:background}

\subsection{Problem Formulation and Related Work}   \label{subsec:related_work}

\paragraph{Problem formulation.}
For each task \(d\) in the set of evaluated tasks \(\mathcal{D}\), an example is an image-text pair \(x=(v,t)\), where \(v\) is the meme image and \(t\) contains the task instruction and any supplied OCR; the gold label \(y\) lies in the task's label set \(\mathcal{C}_d\). A frozen LVLM produces a native prediction \(\hat{y}^{(d)}_{\mathrm{nat}}(x)\) through constrained decoding or task-specific label scoring. From layer \(\ell\) and token role \(s\), we extract a representation \(\phi_{\ell,s}(x)\), defined in Section~\ref{subsec:sparse}, and fit a supervised probe using only the task training split. On the evaluation split, we define the empirical readout gap as
\begin{equation}
    \widehat{\Delta}_{\ell,s,d} =
    \operatorname{MacroF1} \!\left(\hat{\mathbf{y}}_{\mathrm{probe},\ell,s,d},\mathbf{y}_{d}\right)
    -
    \operatorname{MacroF1} \!\left(\hat{\mathbf{y}}_{\mathrm{nat},d},\mathbf{y}_{d}\right)
    \label{eq:readout-gap}
\end{equation}

where \(\mathbf{y}_{d}\) is the vector of gold reporting labels, \(\hat{\mathbf{y}}_{\mathrm{probe},\ell,s,d}\) and \(\hat{\mathbf{y}}_{\mathrm{nat},d}\) are the probe and native predictions on the same examples, and \(\operatorname{MacroF1}\) is the unweighted mean of per-class F1 scores. A positive \(\widehat{\Delta}_{\ell,s,d}\) means that the selected probe can access label-relevant information beyond what is expressed by the native output. However, it does not establish model belief, identify a unique causal mechanism, or constitute a fair comparison between equally trained classifiers~\citep{hewitt2019designing,belinkov-2022-probing}. We therefore separate three questions. \emph{Decodability} asks whether the label can be read from the representation, \emph{routing} asks whether the corresponding directions influence the native output, and \emph{recoverability} asks whether the signal can be used to improve the model's own predictions.

\paragraph{Harmful meme detection.}
Harmful memes are compositional objects whose meaning can depend on the juxtaposition of visual entities, overlaid text, sarcasm, cultural references, and the identity of the targeted group. The Hateful Memes Challenge made this property explicit through benign confounders, where altering either the image or the text can reverse the label while preserving the other modality~\citep{kiela2020hateful}. This construction penalizes unimodal priors, although strong multimodal performance alone is not sufficient to demonstrate that a model has learned a genuine cross-modal interaction~\citep{hessel-lee-2020-multimodal}. Prior work on harmful meme detection emphasizes stronger multimodal fusion/alignment, uncertainty training, and LVLM prompting/context/rationale supervision, but typically focuses on optimizing the final classifier without disentangling representational failures from readout failures in LVLMs~\citep{kumar-nandakumar-2022-hate,10.1145/3701716.3718382,yang-etal-2024-uncertainty-guided,huang-etal-2025-evolver,10.1609/aaai.v40i45.41220,kmainasi-etal-2025-memeintel}.

\paragraph{Probing, sparse representations, and readout gaps.}
Diagnostic probes test which properties are recoverable from frozen states, while sparse autoencoders (SAEs) provide a high-dimensional, sparsely active coordinate system for analyzing those states~\citep{hewitt2019designing,belinkov-2022-probing,ICLR2024_1fa1ab11,marks2025sparse}. Prior work has identified structured multimodal features, failure-predictive internal states, and discrepancies between internal representations and emitted answers in settings involving CLIP, LVLM hallucination, and language-model truthfulness~\citep{goh2021multimodal,kogilathota-etal-2026-halp,orgad2025llmsknowshowintrinsic}. We instead ask whether role-conditioned directions that discriminate harmful-content labels are the same directions that control the LVLM's native label output. We compare dense, base-SAE, residual-SAE, and cross-layer readouts, and reserve claims about native model use for output-alignment tests and causal interventions. Because our probes are supervised and harmfulness labels depend partly on social conventions, we use \emph{readout gap} narrowly to mean task label accessibility, not latent belief or knowledge.

\paragraph{Steering, output control, and distillation.}
Hidden-state steering and decoding-time control can make internal signals behaviorally effective, while knowledge distillation and low-rank adaptation can absorb an external readout into model parameters~\citep{li2023inference,yang-klein-2021-fudge,hinton2015distilling,hu2021lora}; recent work distills intermediate-layer probe predictions instead of teacher outputs and re-injects probe signals into decoding~\citep{brown2026taskspecific,quang2026senses}. We evaluate residual reconstruction, calibrated direct logit routing, and jointly supervised LoRA as three complementary recovery paths. We also compare dedicated and shared adapters because heterogeneous multi-task adaptation can produce negative transfer~\citep{standley2020tasks}. See Appendix~\ref{app:expanded_related_work} for an expanded review of related work.

\subsection{Sparse Representations, Token Roles and Readout Mechanisms}    \label{subsec:sparse}

Let \(h_{\ell,t}(x)\in\mathbb{R}^{d_\ell}\) denote the residual-stream state produced by the frozen LVLM at layer \(\ell\) and token position \(t\). SAEs provide a sparse coordinate system for these states~\citep{ICLR2024_1fa1ab11,ICLR2025_42ef3308}. A frozen public/base SAE is applied first, and a second SAE is trained on its normalized reconstruction residual~\citep{koriagin2025teach,poduval2026resae}:
\begin{equation}
    \begin{aligned}
    z^b_{\ell,t} &= E_b(h_{\ell,t}),
    &
    \widehat{h}^{\,b}_{\ell,t} &= D_b(z^b_{\ell,t}),
    &
    r_{\ell,t} &= h_{\ell,t}-\widehat{h}^{\,b}_{\ell,t},
    \\
    z^r_{\ell,t} &= \operatorname{JumpReLU} \!\left(E_r\!\left(N(r_{\ell,t})\right)\right),
    &
    \widehat{r}_{\ell,t} &= N^{-1}\!\left(D_r(z^r_{\ell,t})\right),
    &
    \widehat{h}^{\,\mathrm{joint}}_{\ell,t} &= \widehat{h}^{\,b}_{\ell,t}+\widehat{r}_{\ell,t}.
    \end{aligned}
    \label{eq:residual-sae}
\end{equation}

Here, \(E_b,D_b\) and \(E_r,D_r\) are the encoders and decoders of the base and residual SAEs, \(z^b_{\ell,t}\) and \(z^r_{\ell,t}\) are their sparse codes, \(\widehat{h}^{\,b}_{\ell,t}\) is the base reconstruction, \(r_{\ell,t}\) is its reconstruction residual, \(\widehat{r}_{\ell,t}\) is the decoded residual, and \(d_\ell\) is the hidden width of layer \(\ell\). \(N\) is a per-dimension mean and standard-deviation normalization whose statistics come from the residual-SAE training data, \(N^{-1}\) undoes it, and \(\operatorname{JumpReLU}\) zeroes pre-activations below a learned per-latent threshold~\citep{rajamanoharan2024jumping}. The public/base dictionary remains frozen and only \((E_r, D_r)\) is trained. We evaluate the base and residual sparse codes as alternative probe representations, while \(\widehat{h}^{\,\mathrm{joint}}_{\ell,t}\) is used by the residual reconstruction hook. Figure~\ref{fig:residual_sae} summarizes the representation construction in panel A and the downstream evaluation and recovery paths in panel B. Objectives, checkpoint source, and complete readout definitions are given in Appendix~\ref{app:representation-details}.

\paragraph{Sparse dictionaries.}
For Gemma, we use public dictionaries trained on instruction-tuned activations and, where selected on calibration data, sparse crosscoders that map several layers into a shared code with layer-specific decoders~\citep{lieberum-etal-2024-gemma,lindsey2024crosscoders}. For Qwen, the six headline analyses use the same public layer-20 SAE trained on base-model activations \citep{deng2026qwenscopeturningsparsefeatures}; we retain a separately trained residual SAE as an ablation. Here, \emph{base SAE} denotes the first dictionary in Equation~\ref{eq:residual-sae}, whereas \emph{base-model SAE} describes the training regime of the Qwen checkpoint.\footnote{The uneven Gemma and Qwen SAE regimes reflect public-checkpoint availability. No instruction-tuned SAE matched the selected Qwen3.5-9B backbone; therefore, our controlled comparisons are the within-family native--hook--probe differences, and we interpret cross-family agreement with Gemma-3 only as replication of the readout-gap phenomenon, not as a model or SAE leaderboard.}


\paragraph{Token roles.}
We partition positions by their origin into prompt/OCR positions
\(\mathcal{T}_{\mathrm{prompt}}\), image-origin positions \(\mathcal{T}_{\mathrm{image}}\), and autoregressively generated positions \(\mathcal{T}_{\mathrm{gen}}\). We additionally use \(\mathcal{T}_{\mathrm{pre}} = \mathcal{T}_{\mathrm{prompt}}\cup\mathcal{T}_{\mathrm{image}}\)
and
\(\mathcal{T}_{\mathrm{all}} = \mathcal{T}_{\mathrm{pre}}\cup\mathcal{T}_{\mathrm{gen}}\).

These are source labels, not modality-pure states. Contextual processing can place visual evidence in prompt-position states and linguistic evidence in image-position states. Given the token-level feature vector \(z_{\ell,t}(x)\) from the selected sparse dictionary, we construct one representation per meme and role by feature-wise max pooling:
\begin{equation}
    \bigl[\phi_{\ell,s}(x)\bigr]_j = \max_{t\in\mathcal{T}_s(x)}\bigl[z_{\ell,t}(x)\bigr]_j,
    \qquad
    s\in \{\mathrm{prompt},\mathrm{image},\mathrm{gen}, \mathrm{pre},\mathrm{all}\}.
    \label{eq:role-pooling}
\end{equation}
Here \(j\) indexes sparse features and \(\mathcal{T}_s(x)\) is the set of role-\(s\) positions in example \(x\). This records whether feature \(j\) is strongly active anywhere within a role. The pre-generation representation tests accessibility before decoding, generated-token states describe computation after decoding has begun, and the all-token representation captures the full trajectory. A strong generated or all-token probe thus does not show that the same information was available before the answer was generated. Dense controls apply the same pooling to the role-conditioned hidden states instead of the SAE codes.

\paragraph{Readout pathways.}
Starting from the same frozen LVLM, we compare four paths. The \emph{native} path uses the model's constrained label scores. The \emph{sparse probe} applies a supervised readout to \(\phi_{\ell,s}(x)\) and measures label decodability. The \emph{residual reconstruction hook} blends \(\widehat{h}^{\,\mathrm{joint}}_{\ell,t}\) into the residual stream and then continues through the model's original upper layers, testing whether reconstruction alone makes the signal usable by the native pathway. Finally, \emph{direct logit routing} adds a calibrated probe score directly to the output layer; the correction skips the upper layers, although the native forward pass is still needed. Section~\ref{sec:latent_signal} defines this router and the subsequent jointly supervised LoRA procedure. Together, these paths distinguish information that is accessible, information that survives the model's native routing, and information that can be converted into improved model behavior.

\subsection{Datasets and Evaluation Setup}      \label{subsec:datasets}

We evaluate six primary binary multimodal harmful-content tasks: CrisisHateMM, FHM, MAMI, HarMeme, MMHS150K, and MultiOFF~\citep{bhandari2023crisishatemm,kiela2020hateful,gasparini2022benchmark,pramanick2021detecting,gomez2020exploring,suryawanshi2020multimodal}. Table~\ref{tab:primary_datasets} reports the effective train, validation, and test counts used by the primary binary pipeline. Experiment-specific subdivisions of these splits for calibration and reporting are described in Appendix~\ref{app:data-eval}. For HarMeme, the two harmful classes are merged into \textsc{Harmful}; for MMHS150K, the five hate subtypes are merged into \textsc{Hate}. Their original fine-grained taxonomies and CrisisHateMM target attribution are analyzed separately. Multilingual robustness uses fixed internal holdouts from EXIST-2025 ($500$ examples, equally divided between English and Spanish) and MultiBully ($1000$ Hindi--English examples)~\citep{10.1007/978-3-032-04354-2_16,10.1145/3477495.3531925}; neither contributes to the six-task mean score.


\begin{table}[ht]
    \centering
    \caption{\textbf{Primary binary tasks and effective split sizes.} Counts are usable examples after fixed preprocessing and before any experiment-specific calibration/reporting sub-splits. Boldface identifies the locked reporting split.}
    \label{tab:primary_datasets}
    \small
    \renewcommand{\arraystretch}{0.96}
    \begin{tabular}{@{}ccccc@{}}
        \toprule
        Dataset & Binary target & Train & Validation & Test \\
        \midrule
        CrisisHateMM & hate / no hate & 3,600 & \textbf{443}\tablefootnote{The released CrisisHateMM test directory does not provide usable gold labels. We therefore treat its labeled validation set as the final evaluation split and do not use it for fitting the reported probe.} & -- \\
        FHM & hateful / not hateful & 7,938 & 469 & \textbf{1,000} \\
        MAMI & misogynous / not misogynous & 9,000 & 1,000 & \textbf{1,000} \\
        HarMeme & harmful / not harmful & 3,013 & 177 & \textbf{354} \\
        MMHS150K & Hate / NotHate & 134,823 & 5,000 & \textbf{10,000} \\
        MultiOFF & offensive / non-offensive & 445 & 149 & \textbf{149} \\
        \bottomrule
    \end{tabular}
\end{table}

All trainable components use task training data only. Layers, token roles, representations, checkpoints, thresholds, and routing coefficients are selected on training or disjoint calibration data and frozen before reporting. Native predictions use constrained task-specific label scoring; originally multiclass datasets are scored in their full label space before the fixed binary evaluation. Macro-F1 is the primary metric, and all systems within a model family are compared on identical example IDs. Full dataset construction, split rules, decoding details, secondary metrics, and comparison scope are provided in Appendix~\ref{app:data-eval}.

\section{Internal Representations versus Generative Head}       \label{sec:internal_repr}

\paragraph{Broad Evaluation Comparison.}
Table~\ref{tab:broad-comparison} compares locked macro-F1 for the native LVLM, the selected sparse readout, and, for Qwen, the residual reconstruction hook. Sparse readouts outperform native prediction on every task in both model families. Gemma improves from \(0.532\) to \(0.714\) mean macro-F1 (\(+0.182\)), while Qwen improves from \(0.432\) to \(0.740\) (\(+0.308\)). Qwen provides cleaner cross-task control because all six tasks use the same public layer-20 base SAE and linear readout family. Gemma's native-probe gaps remain controlled within tasks, but its mean combines calibration-selected single-layer, residual, and cross-layer readouts. We therefore interpret it as a heterogeneous replication rather than a matched Gemma-Qwen comparison.




The Qwen hook improves five tasks and raises the mean to \(0.486\), but recovers only \(17.6\%\) of the native-to-probe gap. Sparse reconstruction helps the native output somewhat, but most of the accessible signal still does not reach it. As defined in Section~\ref{subsec:related_work}, these gains measure supervised accessibility; the probe is trained while the native classifier is not, so the comparison is not like-for-like. Per-task analysis, base-versus-residual SAE comparisons, and fine-grained results are reported in Appendices~\ref{app:broad-comparison-details} and~\ref{app:fine-grained}.




\begin{table}[ht]
    \centering
    \caption{\textbf{Macro-F1 scores on the six primary binary tasks.} Gemma uses its calibration-selected sparse readout while Qwen uses the same public layer-20 base-SAE probe across tasks. The hook modifies Qwen's native pathway through residual reconstruction. Means are unweighted across tasks.}
    \label{tab:broad-comparison}
    \small
    \setlength{\tabcolsep}{4.5pt}
    \renewcommand{\arraystretch}{0.96}
    \begin{tabular}{@{}lccccc@{}}
        \toprule
        & \multicolumn{2}{c}{Gemma-3-4B-IT}
        & \multicolumn{3}{c}{Qwen3.5-9B-Base} \\
        \cmidrule(lr){2-3}
        \cmidrule(lr){4-6}
        Dataset & Native & Sparse & Native & Hook & SAE \\
        \midrule
        CrisisHateMM & 0.606 & 0.819 & 0.310 & 0.349 & 0.860 \\
        FHM          & 0.648 & 0.703 & 0.637 & 0.692 & 0.712 \\
        MAMI         & 0.409 & 0.731 & 0.408 & 0.521 & 0.755 \\
        HarMeme      & 0.719 & 0.776 & 0.394 & 0.394 & 0.810 \\
        MMHS150K     & 0.292 & 0.630 & 0.524 & 0.556 & 0.587 \\
        MultiOFF     & 0.519 & 0.623 & 0.320 & 0.404 & 0.716 \\
        \midrule
        \textbf{Mean} & \textbf{0.532} & \textbf{0.714} & \textbf{0.432} & \textbf{0.486} & \textbf{0.740} \\
        \bottomrule
    \end{tabular}
\end{table}

\paragraph{Where is the Signal? Token-Role and Representation Ablations.}
Using the calibration-only protocol of Section~\ref{subsec:datasets}, role-conditioned probes reveal no universal position of harmful content information. On Gemma, image-position features outperform generated features on CrisisHateMM and HarMeme, whereas generated features are stronger on MAMI, MMHS150K, and MultiOFF; FHM is nearly tied. The largest shift occurs on MMHS150K, where moving from image to generated states raises macro-F1 from \(0.531\) to \(0.617\) for the binary task and from \(0.232\) to \(0.468\) for the six-class task. MAMI instead peaks at prompt/OCR states (\(0.845\)). Because generated states are observed only after decoding begins, they indicate where the computation becomes linearly separable, not necessarily what was available before the decision. Even so, image or pre-generation readouts still exhibit large readout gaps on CrisisHateMM and HarMeme, while the selected FHM readout uses prompt--image cross-layer features. The signal is already accessible before generation, and decoding can make it more separable.

Sparse representation is similarly task-dependent. Relative to probes over corresponding dense hidden states, it helps substantially on CrisisHateMM (\(+0.283\)), modestly on FHM and six-class MMHS150K, and negligibly on MAMI, HarMeme, and MultiOFF; random sparse controls remain near chance. Most of the harmful content signal is thus already present in the dense states. The SAE makes it easier to see and supplies the feature basis needed for the alignment and intervention tests in Section~\ref{subsec:mechanism}. FHM additionally resists single-role and simple pooled interaction readouts, motivating the pair-aware low-rank analysis in Section~\ref{subsec:fhm_bilinear}. Full role sweeps, dense-state controls, and FHM ablations are reported in Appendix~\ref{app:role-representation-ablation}.

\section{Mechanistic Structure of Readout Gap}      \label{sec:mech_structure}

\begin{figure}[ht]
    \centering
    \includegraphics[width=0.75\linewidth]{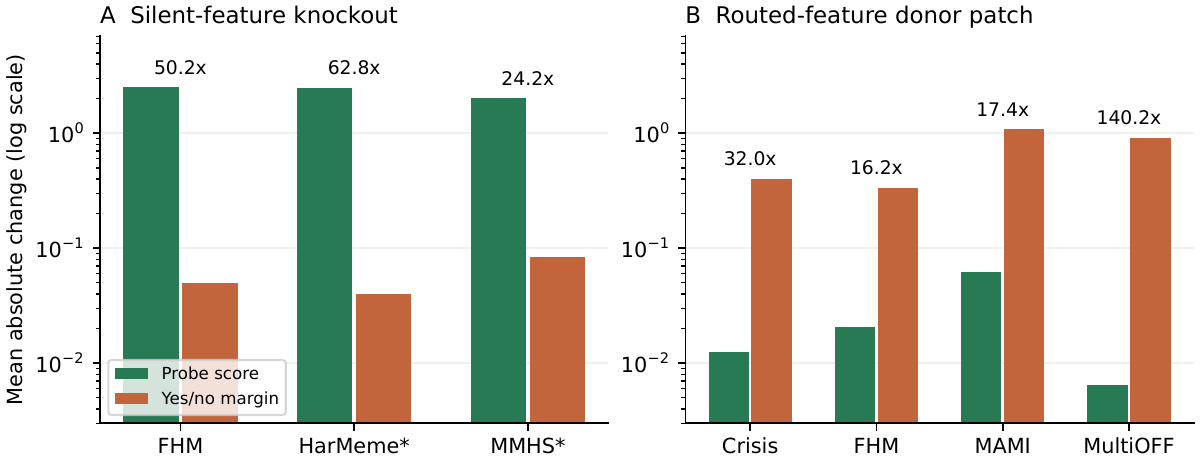}
    \caption{\textbf{Decodability and native-output control dissociate in Qwen.} \textbf{A. Silent-feature knockout.} On up to \(N_{\mathrm{KO}}=15\) positive-class, probe-correct, native-wrong examples per reported task, the $20$ task-specific silent features are zeroed at every layer-20 token position. Bars show the mean absolute paired changes in the recomputed probe score and monitored native margin. \textbf{B. Routed-feature donor patching.} On up to \(N_{\mathrm{pair}}=5\) eligible donor--target pairs per task, the donor's max-over-token activation for each of the $20$ routed features is written to every target token position before the target forward pass continues. Bars again show mean absolute paired changes; labels give the larger-to-smaller sensitivity ratio. Asterisks mark yes/no margins that are diagnostic rather than the task's native label-score decision.}
    \label{fig:alignment-mediation}
\end{figure}


\subsection{Discriminative versus Routed Directions}    \label{subsec:mechanism}
A strong sparse probe need not rely on the directions that control the LVLM's own answer. Let \(d_j\) be the decoder direction of SAE feature \(j\) (not to be confused with the task index \(d\)), and let \(w_{d,j}\) be its task-\(d\) probe coefficient, oriented so that positive scores favor the harmful class. We compare this discriminative importance with direct access to the yes/no output channel:
\begin{equation}
    \begin{aligned}
    u_{\mathrm{yn}} &= W_U^{f\top} \left(e_{\mathrm{yes}}-e_{\mathrm{no}}\right),
    &
    a_j &= d_j^\top u_{\mathrm{yn}},
    \\
    \mathcal{S}_d &= \left\{j\in\operatorname{TopK}_k\!\left(|w_{d,\cdot}|\right): |a_j|<\tau\right\},
    &
    \mathcal{R}_{\mathrm{yn}} &= \operatorname{TopK}_k\!\left(|a_\cdot|\right)
    \end{aligned}
    \label{eq:discriminative-routed-sets}
\end{equation}

where \(W_U^f\) is the effective unembedding (output head) with final normalization folded into it, \(e_{\mathrm{yes}}\) and \(e_{\mathrm{no}}\) are one-hot vocabulary vectors for the \texttt{yes} and \texttt{no} tokens, \(a_j\) is feature \(j\)'s direct effect on the yes-minus-no logit margin, \(\operatorname{TopK}_k(\cdot)\) returns the indices of the \(k\) largest entries, \(\tau=0.10\) is the output-silence threshold, and \(k=20\). The task-specific set \(\mathcal{S}_d\) contains probe-important but output-silent features, whereas \(\mathcal{R}_{\mathrm{yn}}\) contains features selected only for direct access to the native yes/no channel. The yes/no anchor is the native decision channel only for literal yes/no tasks and a diagnostic for full-label-scored tasks, and \(a_j\) ignores the remaining transformer blocks, so it is not a full causal effect~\citep{elhage2021mathematical,belrose2023eliciting}.

For Qwen's public base SAE, no analyzed feature combines high probe importance with high direct output alignment on five of six tasks (MAMI: \(1\%\) of records). Figure~\ref{fig:alignment-mediation} shows the causal pattern. On the raw score scales, knocking out silent features changes the probe score \(24.2\)--\(62.8\) times more than the monitored yes/no margin, whereas donor-patching routed features change the native margin \(16.2\)--\(140.2\) times more than the probe across the four tasks with matched literal yes/no decoding. These ratios depend on the two score scales and are not scale-invariant mediation fractions. Example-level bootstrap intervals support the knockout direction (FHM: \(\Delta\)probe \(=-2.512\;[-2.769,-2.250]\) versus \(\Delta\)logit \(=-0.033\;[-0.063,-0.008]\)), whereas the five-pair patch cohorts remain exploratory. Routed interventions establish output control rather than improved correctness~\citep{vig2020investigating,meng2022locating,marks2025sparse}.

Gemma replicates the static and probe-side separation: its routed and probe-derived feature sets have zero overlap on all four analyzed tasks, and Jacobian-based~\citep{gurnee2026verbalizable} sensitivity places the strongest probe directions outside the dominant output-routing set. Its native-output mediation is less uniform, so we claim no universal orthogonal decomposition; conclusions are local to the analyzed layers, dictionaries, and output anchors (Appendix~\ref{app:causal-details}).

\begin{figure}[t]
    \centering
    \includegraphics[width=0.95\linewidth]{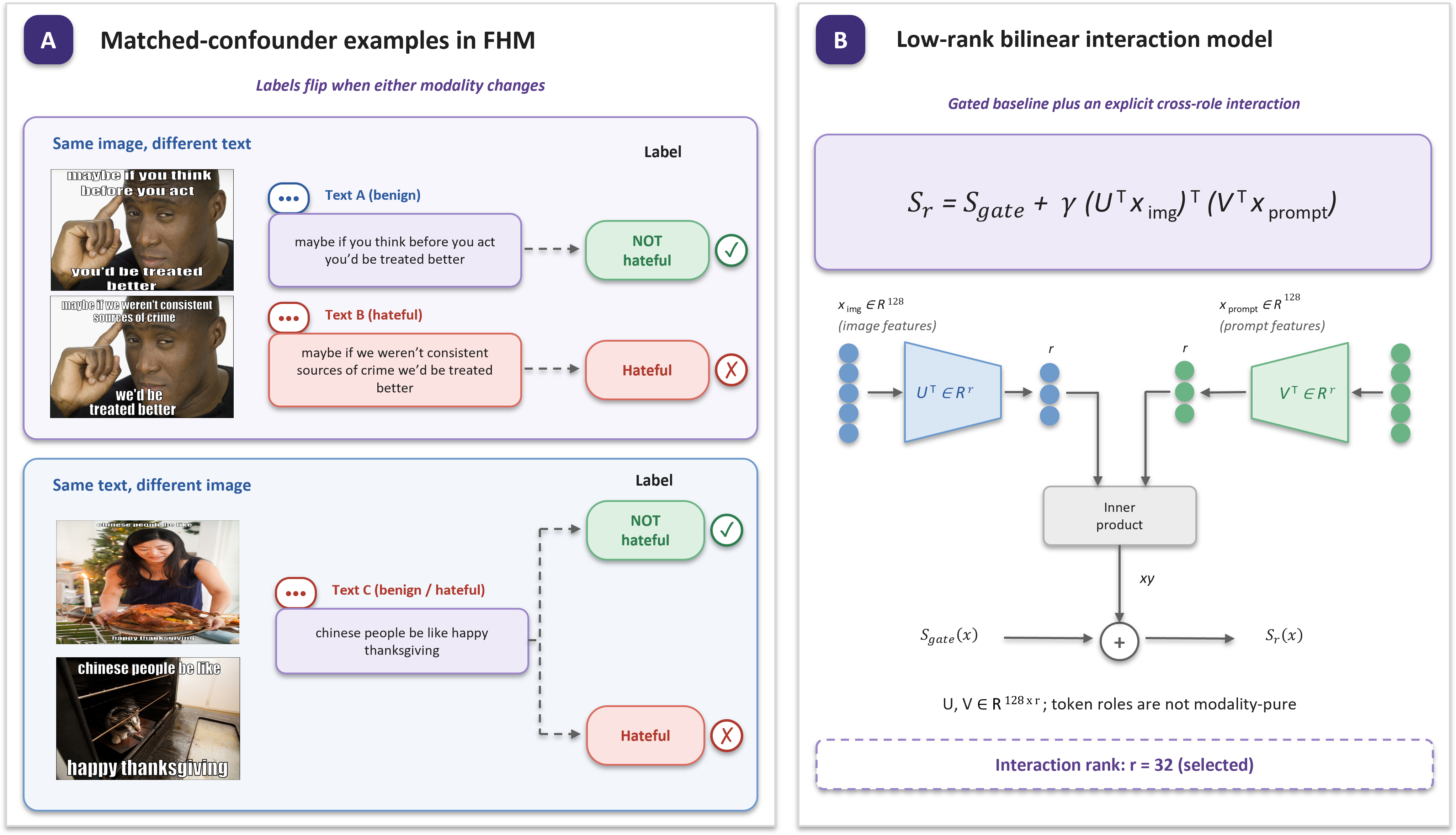}
    \caption{\textbf{Confounders and the bilinear readout.} \textbf{A.} Keeping the background image or accompanying text constant can reverse the gold label when the other component changes; symbols indicate gold classes, not model correctness. \textbf{B.} Two role-conditioned projections form an inner product, scaled by $\gamma$ and added to the gated baseline. Image and prompt denote token origins, not modality-pure information.}
    \label{fig:fhm-low-rank}
\end{figure}

\subsection{Low-Rank Cross-Modal Structure on FHM}      \label{subsec:fhm_bilinear}

Because this Gemma-3-12B experiment changes both model scale and readout family, we exclude it from the six-task mean in Section~\ref{sec:internal_repr}. FHM's benign confounders can reverse the label while holding the image or text fixed (Figure~\ref{fig:fhm-low-rank}A), so combining two role-specific predictors does not establish a cross-modal interaction~\citep{kiela2020hateful,hessel-lee-2020-multimodal}. From the layer-31 sparse representation, we construct standardized max-pooled image-position and prompt/OCR vectors \(x_{\mathrm{img}},x_{\mathrm{prompt}}\in\mathbb{R}^{128}\) using features selected by a train-only confounder audit. We augment the pair-aware gated baseline \(s_{\mathrm{gate}}\) with an explicit rank-\(r\) interaction:
\begin{equation}
    \begin{aligned}
    s_r(x) &= s_{\mathrm{gate}}(x) + \gamma x_{\mathrm{img}}^\top U V^\top x_{\mathrm{prompt}}
    \\
    &= s_{\mathrm{gate}}(x) + \gamma \sum_{q=1}^{r}\left(u_q^\top x_{\mathrm{img}}\right)\left(v_q^\top x_{\mathrm{prompt}}\right),
    \qquad
    U,V\in\mathbb{R}^{128\times r}.
    \end{aligned}
    \label{eq:fhm-low-rank}
\end{equation}

Here \(s_{\mathrm{gate}}\) is the gated pairwise score defined in Appendix~\ref{app:fhm-low-rank-details}, \(U\) and \(V\) are learned factor matrices with columns \(u_q\) and \(v_q\), \(r\) is the interaction rank, and \(\gamma\) is a learned scalar weight on the interaction term (Figure~\ref{fig:fhm-low-rank}B). This parameterization uses \(256r\) interaction parameters rather than a full \(128^2\) matrix \citep{kim2016hadamard}. Training adds ranking losses on opposite-label same-image and same-text pairs to sample-level classification, and all choices are frozen before the locked test evaluation.

\begin{figure}[ht]
    \centering
    \includegraphics[width=0.90\linewidth]{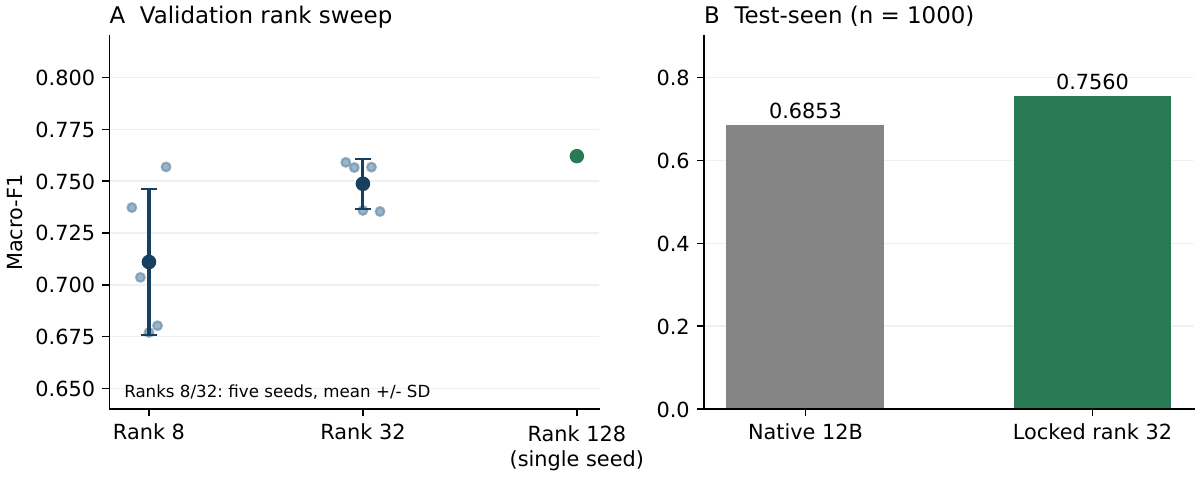}
    \caption{\textbf{FHM rank sweep and locked test.} \textbf{A.} Validation macro-F1 ($n=375$); dots are seeds, bars SD. \textbf{B.} Test-seen macro-F1 ($n=1{,}000$), pair-aware readout versus native Gemma-3-12B.}
    \label{fig:fhm-quantitative}
\end{figure}

Figure~\ref{fig:fhm-quantitative}A shows that rank 8 is unstable across seeds, whereas rank 32 is stable and approaches the single-seed rank-128 result. The calibration-selected rank-32 checkpoint reaches \(0.7560\) test-seen macro-F1 versus \(0.6853\) for native Gemma-3-12B (Figure~\ref{fig:fhm-quantitative}B), a gain of \(0.0707\) (paired bootstrap 95\% CI \([0.0461,0.0954]\)) for the complete pair-aware readout, not the bilinear term alone.

No single feature pair accounts for the interaction: $11$ of the $32$ factors have measurable leave-one-out effects, but removing any one changes validation macro-F1 by at most \(0.007\), and a within-role quadratic control adds only \(0.004\). Analogous bilinear readouts underperform their simpler controls for Gemma-3-4B and Qwen, so we claim a distributed rank-32 image-by-prompt interaction in the analyzed Gemma-3-12B layer-31 representation, not a universal FHM mechanism (Appendix~\ref{app:fhm-low-rank-details}).

\section{Recovering and Stress-Testing the Latent Signal}       \label{sec:latent_signal}

\paragraph{Direct Logit Routing and Probe Distillation.}
Section~\ref{subsec:mechanism} suggests two recovery paths: bypass the native routing bottleneck at the output, or absorb the readout into the model. For task \(d\), let \(m_d(x)\) be the signed native decision margin (positive favors harmful) and \(u_d(x)\) the frozen probe's decision score. We use the calibration-only router
\begin{equation}
    m^{\mathrm{route}}_d(x) = m_d(x) + \beta_d\frac{u_d(x)-\mu_d}{\sigma_d+\epsilon}
    \label{eq:direct-logit-routing}
\end{equation}

where the probe-score mean \(\mu_d\), standard deviation \(\sigma_d\), and mixing coefficient \(\beta_d\geq0\) are fitted on a disjoint calibration partition and frozen, and \(\epsilon\) is a small stability constant.

Across all six tasks, native-only threshold calibration raises mean macro-F1 from \(0.4396\) to \(0.6421\). On the five tasks with complete matched probe-score artifacts, the raw native, calibrated native, probe, and routed means are \(0.4640\), \(0.6593\), \(0.7438\), and \(0.7537\), respectively. Thus calibration explains \(69.8\%\) of the raw native-to-probe gap, but routing adds \(0.0944\) beyond calibrated native scoring and slightly exceeds the probe mean. Native AUROC is only \(0.579\) on MMHS150K and \(0.661\) on MultiOFF, so these tasks also have ranking failures that no threshold can repair: the diagnosis is output miscalibration plus additional information available to the sparse readout.

To remove the SAE and probe at inference, we train Gemma-3-4B-IT LoRA adapters with gold-label cross-entropy, probe KL supervision, and, where available, matched-pair ranking~\citep{hinton2015distilling,hu2021lora}. On the $1000$ FHM test-seen examples, the dedicated joint objective reaches \(0.7138\) macro-F1, versus \(0.6482\) for the base model and \(0.6462\) for a gold-only adapter, so KL-and-pair supervision helps in this dedicated setting. For the shared seven-task adapter the conclusion reverses: the compute-matched gold-only mean is \(0.5780\) versus \(0.5686\) for the joint objective, which beats gold-only only on FHM, MAMI, and HarMeme and drops most on six-class MMHS150K (\(0.1886\) versus \(0.2688\)). Per-task results appear in Appendix~\ref{app:recovery-details}.

\paragraph{Multilingual and Visio-Textual Robustness.}
We test whether the readout gap is confined to English or explained by the provided OCR via a visual credit audit~\citep{liu2026visualcreditauditmultimodal}. On the fixed EXIST holdouts, Qwen cross-language and bilingual probes reach \(0.7469\)--\(0.8121\) macro-F1; Gemma stays above chance, but its probe-over-native advantage changes sign across split seeds, so for Gemma we claim only cross-lingual decodability. On MultiBully, Qwen's all-token probe reaches \(0.7259\) versus \(0.3401\) natively, and Gemma's residual-SAE probe beats native by \(0.0915\pm0.0226\) across five seeds; silent-feature knockout stays probe-dominant beyond English (Appendix~\ref{app:robustness-details}).


Explicit OCR contributes but does not explain the result: removing it changes the strongest text-bearing readouts by at most \(0.0671\) macro-F1 and the MultiBully image-position probe by only \(0.0011\). In contrast, same-split image permutation reduces probe macro-F1 by \(0.148\)--\(0.206\) and routed macro-F1 by \(0.160\)--\(0.292\) on every audited dataset, and image blanking or shuffling breaks \(55.8\%\)--\(88.8\%\) of originally correct probe or router predictions, versus \(0.0\%\)--\(38.1\%\) for the native classifier. The signal thus holds beyond English and depends heavily on the paired image. These dependence tests do not isolate modality-pure circuits (Appendix~\ref{app:robustness-details};~\citealp{hessel-lee-2020-multimodal}).

\section{Conclusion and Future Work}     \label{sec:conclusion}

Our results separate decodability, native routing, and correctness. Across two LVLM families and six tasks, sparse readouts recover label information the native output expresses weakly, and Qwen interventions separate probe-discriminative from output-routed directions (Figure~\ref{fig:alignment-mediation}). Native threshold calibration explains most of the matched gap, yet direct routing adds \(0.094\) mean macro-F1 beyond it. Joint LoRA supervision helps in some settings, but gold-only controls show the gain is not generally due to the probe targets. Gemma-3-12B also shows a distributed FHM interaction.



\paragraph{Limitations and future work.}
Probes are supervised while the native output is untuned; mechanistic claims are local to the analyzed layers and anchors, with some small or out-of-distribution interventions; generated-token probes are post-generation readouts; multilingual tests use internal holdouts; and fine-grained classes remain hard. Future work should test larger LVLMs, multicultural data, position-matched causal tracing, and subtype-preserving routing.

\bibliography{iclr2027_conference}
\bibliographystyle{iclr2027_conference}

\clearpage
\begingroup
\renewcommand{\contentsname}{Appendix Contents}
\hypersetup{linktoc=all}
\pdfbookmark[0]{Appendix Contents}{appendix-contents}
\tableofcontents
\endgroup
\clearpage

\appendix
\addtocontents{toc}{\protect\setcounter{tocdepth}{2}}
\AddToHook{cmd/section/before}{\FloatBarrier}

\section{Expanded Related Work}
\label{app:expanded_related_work}

\paragraph{Hateful meme detection, localization, explanation, and intervention.}
The harmful-meme literature has expanded beyond binary classification. MOMENTA~\citep{pramanick2021detecting} jointly predicts harmfulness and the targeted social entity, while HateSieve localizes hateful regions and text-image evidence through context-aware contrastive learning~\citep{su-etal-2025-context}. MemeIntel introduces explanation-enhanced resources and jointly optimizes label detection and rationale generation~\citep{kmainasi-etal-2025-memeintel}. These objectives are complementary to mechanistic analysis but answer a different question: generated rationales or localized regions describe a model's claimed evidence, whereas our probes and interventions examine the internal signals that support a decision and whether those signals affect the native output. Hate-CLIPper~\citep{kumar-nandakumar-2022-hate} and context-aware contrastive systems are useful references for task performance. They are not controlled baselines for our mechanistic question, though, because they change the backbone, supervision, and fusion architecture, whereas our native, hook, probe, and routed paths hold the LVLM and examples fixed. For the same reason, we make no leaderboard claim and do not claim that the probe approaches a task-specific optimum.


\paragraph{Domain, language, and cultural variation.}
Harmful meme systems are sensitive to changes in meme format, OCR quality, language, and annotation culture. Models trained on the Hateful Memes Challenge degrade on naturally occurring Pinterest memes whose captions must be recovered through OCR~\citep{kirk-etal-2021-memes}. MUTE supplies Bengali and Bengali-English code-mixed memes~\citep{hossain-etal-2022-mute}, while Multi\(^{3}\)Hate presents parallel memes in five languages with annotations from culturally distinct populations~\citep{bui-etal-2025-multi3hate}. Retrieval-guided and retrieval-augmented methods attempt to improve adaptation to new and evolving examples~\citep{mei-etal-2024-improving,mei-etal-2025-robust}. Our multilingual experiments are evidence about representational transfer under fixed internal holdouts; they do not show that harmfulness has a single language or culture-independent decision boundary.

\paragraph{Interpretability of vision-language representations.}
Early analyses of CLIP identified neurons responding to concepts across visual objects, written words, symbols, and associated entities~\citep{goh2021multimodal}. Later SAE-based approaches learned sparse features in vision encoders and used feature interventions to alter multimodal generation~\citep{pach2026sparse}. SAE-V studies multimodal alignment and data quality through sparse features~\citep{lou2025saevinterpretingmultimodalmodels}, while VL-SAE learns a shared concept set with modality-specific decoders~\citep{shen2025vlsae}. These methods focus primarily on feature coherence, alignment, or controllability. Our analysis instead asks whether task-discriminative sparse directions coincide with the directions through which an autoregressive LVLM communicates a harmful content decision.

\paragraph{Diagnostic probing and latent readout gaps.}
Multimodal probing has been used to examine relational information across visual contexts~\citep{parfenova-etal-2021-probing}, while neuron-level analyses of CLIP identified units responsive to related concepts across objects, rendered text, and visual symbols~\citep{goh2021multimodal}. More recent LVLM work uses internal states to predict hallucination risk before generation, diagnose failures to connect visual references with factual associations, or intervene on hallucination-related activation patterns~\citep{kogilathota-etal-2026-halp,ashok-etal-2025-vlms,wu-etal-2025-sharp}. Related language-model studies show that truth-related or answer-relevant information can be recovered from internal states even when the emitted response is incorrect~\citep{burns2024discoveringlatentknowledgelanguage,belrose2023eliciting,orgad2025llmsknowshowintrinsic}. These results motivate separating representational availability from external behavior. Concurrent work finds the same separation outside harmful-content moderation. In Gemma-3 and Qwen3 clinical triage, medical SAE features fire on the shared case narrative but are silent at the multiple-choice decision token~\citep{fraile2026internal}. In omnimodal LLMs, hidden states encode premise-perception conflicts that the models rarely reject in their outputs, and a probe-guided logit adjustment partly recovers the behavior~\citep{quang2026senses}. Our work tests the analogous question for multimodal safety labels, separates discriminative from output-routed directions causally, and measures how much of the gap native threshold calibration alone explains. Our setting is narrower because the probes are supervised and harmfulness labels are socially defined, so we test benchmark-label accessibility and separately measure whether the identified directions affect the native output.

\paragraph{Validity of sparse explanations.}
Neither sparse reconstruction nor a human-readable feature description guarantees causal relevance. Probe coefficients can reflect correlated features, SAE dictionaries can split one concept across multiple latents, and feature identities can vary with architecture, width, training data, or initialization. We treat sparse features as a coordinate system for testing hypotheses, not as a complete inventory of what the model computes. Raw-state controls test whether sparsification improves accessibility, base-versus-residual comparisons test dependence on the dictionary construction, and ablation or patching tests whether selected coordinates are functionally load-bearing. This follows the broader move from descriptive feature visualization toward sparse causal circuits~\citep{marks2025sparse}. Our residual SAE follows prior residual-dictionary constructions: boosting trains a secondary SAE on a pretrained SAE's reconstruction error to capture missed domain-specific features~\citep{koriagin2025teach}, and ReSAE residualizes across layers to reduce redundancy under multi-layer interventions~\citep{poduval2026resae}. We use a single-layer, within-dictionary residual and treat it as an ablation rather than as the primary interpretability object.

\paragraph{Alternative routes from representation to behavior.}
Representation engineering methods modify hidden states directly, while decoding-time methods modify output probabilities without changing the underlying computation~\citep{li2023inference,yang-klein-2021-fudge,liu-etal-2021-dexperts}. In LVLMs, SHARP steers dense hallucination-related representations~\citep{wu-etal-2025-sharp}, whereas SAE-based hallucination mitigation intervenes on sparse latent directions~\citep{hua-etal-2025-steering}. Our experiments include both a hidden-state reconstruction path and a logit-space path; the latter is closely related to probe-guided logit adjustment~\citep{quang2026senses}, but we calibrate it on disjoint data and compare it with native-only threshold calibration. The subsequent LoRA experiment asks a stricter question: whether the external readout can be internalized so that neither the SAE nor the probe is required during inference. Using intermediate-layer probe predictions as distillation targets can outperform output-logit distillation on reasoning tasks~\citep{brown2026taskspecific}. Our gold-only controls show that in harmful meme classification this benefit is task-dependent.

\section{Sparse-Representation and Readout Details}
\label{app:representation-details}

\paragraph{SAE objectives.}
For a token-level target \(u_{\ell,t}\), an SAE encoder-decoder pair produces \[z_{\ell,t}=E(u_{\ell,t}), \qquad \widehat{u}_{\ell,t}=D(z_{\ell,t}) \]
and uses a reconstruction--sparsity objective of the form
\begin{equation}
    \mathcal{L}_{\mathrm{SAE}} = \mathbb{E}_{x,t}
    \left[\left\|u_{\ell,t}(x)-D(E(u_{\ell,t}(x)))\right\|_2^2 + \lambda\left\|E(u_{\ell,t}(x))\right\|_0\right].
    \label{eq:appendix-sae-objective}
\end{equation}

Equation~\ref{eq:appendix-sae-objective} is a schematic reconstruction--sparsity objective, where \(z_{\ell,t}\) is the sparse code, \(\widehat{u}_{\ell,t}\) is the reconstruction, \(\|\cdot\|_0\) counts nonzero latents, \(\lambda>0\) weights the sparsity penalty, and the expectation runs over training examples \(x\) and token positions \(t\). The residual-SAE implementation uses variance-normalized reconstruction error (FVU)~\citep{bricken2023towards,ICLR2024_1fa1ab11} and a JumpReLU~\citep{rajamanoharan2024jumping} sparsity penalty. The logged coefficients apply to that normalized objective, not to an unnormalized squared-error loss. For a standard dictionary, \(u_{\ell,t}=h_{\ell,t}\). For the residual SAE trained in this work, \(u_{\ell,t}=N(r_{\ell,t})\), where \(r_{\ell,t}=h_{\ell,t}-\widehat{h}^{\,b}_{\ell,t}\). The decoded residual is mapped back to the original activation scale with \(N^{-1}\). The base SAE remains frozen throughout residual SAE training.

\begin{figure}[ht]
    \centering
    \includegraphics[width=0.95\linewidth]{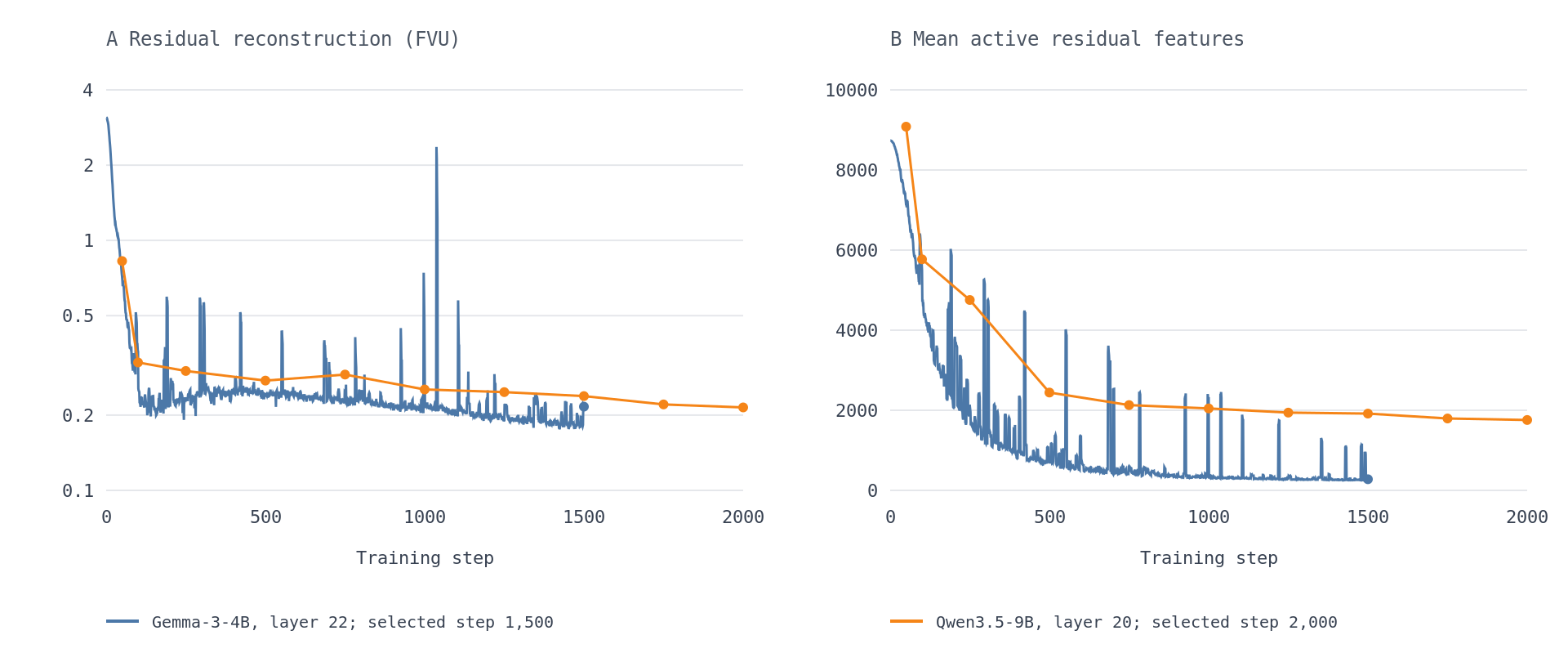}
    \caption{\textbf{Selected residual-SAE training diagnostics.} Training FVU (log scale, left) and mean active features $L_0$ (right) for Gemma-3-4B layer 22 ($\lambda=0.0003$, selected step 1,500) and Qwen3.5-9B layer 20 ($\lambda=0.0001$, selected step 2,000). Reconstruction metrics were not used as a substitute for downstream checkpoint evaluation: better SAE-objective values did not monotonically improve downstream behavior.}
  \label{fig:sae-l0-curves}
\end{figure}

\paragraph{Checkpoint sources.}
The Gemma dictionaries are derived from instruction-tuned Gemma activations and include both single-layer SAEs and sparse crosscoders~\citep{lieberum-etal-2024-gemma,lindsey2024crosscoders}\footnote{\url{https://huggingface.co/google/gemma-scope-2}}. The Qwen headline analyses use a public SAE trained on base-model activations~\citep{deng2026qwenscopeturningsparsefeatures}\footnote{\url{https://huggingface.co/collections/Qwen/qwen-scope}}. 

\paragraph{Dictionary and probe hyperparameters.}
The Gemma-3-4B public layer-22 base SAE has width $65{,}536$. Its selected residual JumpReLU SAE uses $d_{\mathrm{in}}=2{,}560$, expansion factor $8$ ($20{,}480$ latents), $\lambda_{L_0}=3\times10^{-4}$, and checkpoint step $1{,}500$. The Qwen public layer-20 TopK SAE has width $65{,}536$ and $k=100$. Its residual JumpReLU ablation uses $d_{\mathrm{in}}=4{,}096$, expansion factor $8$ ($32{,}768$ latents), $\lambda_{L_0}=10^{-4}$, and checkpoint step $2{,}000$. Both residual SAEs use per-dimension residual normalization, AdamW with learning rate $5\times10^{-5}$ and zero weight decay, 100 warmup steps, batch size $4{,}096$, and gradient-norm clipping at $1.0$.

Unless a nonlinear readout is explicitly named, probes are $L_2$-regularized logistic regressions with inverse regularization strength $C=1.0$ and balanced class weights. Gemma probes use scikit-learn's \texttt{saga} solver for up to $5{,}000$ iterations; tasks with more than $20{,}000$ training examples switch to \texttt{liblinear}, with one-vs-rest wrapping for multiclass targets. Qwen probes first fit a train-only MaxAbs scaler and then use \texttt{lbfgs} for up to $1{,}000$ iterations. These solvers do not expose a user-set learning rate; optimization tolerance otherwise uses the scikit-learn default. Per-task thresholds and any stated regularization overrides are selected only on training or calibration data.


\paragraph{Cross-layer sparse representations.}
For selected Gemma experiments, a sparse crosscoder constructs one shared token-level code from activations at layers \(\mathcal{L}\):
\begin{equation}
    z^{\mathrm{cc}}_t = \rho\!\left(\sum_{\ell\in\mathcal{L}} W^{(\ell)}_{\mathrm{enc}}h_{\ell,t}
        + b_{\mathrm{enc}}\right),
    \qquad
    \widehat{h}_{\ell,t} = W^{(\ell)}_{\mathrm{dec}}z^{\mathrm{cc}}_t + b^{(\ell)}_{\mathrm{dec}}
    \label{eq:appendix-crosscoder}
\end{equation}

where \(\rho\) is the checkpoint-specific sparse nonlinearity, \(W^{(\ell)}_{\mathrm{enc}}\) and \(W^{(\ell)}_{\mathrm{dec}}\) are the layer-specific encoder and decoder matrices, \(b_{\mathrm{enc}}\) is the shared encoder bias, \(b^{(\ell)}_{\mathrm{dec}}\) is the layer-specific decoder bias, and \(\widehat{h}_{\ell,t}\) is the layer-\(\ell\) reconstruction. The shared code tests whether harmful content information is spread across model depth or confined to one analyzed layer.

\paragraph{Native and external readouts.}
For class \(c\), let \(\nu(c)\) denote its textual verbalization, with \(q\)-th token \(\nu(c)_q\) and preceding tokens \(\nu(c)_{<q}\). With \(p_\theta\) the next-token distribution of the frozen LVLM with parameters \(\theta\), the native label score is
\begin{equation}
    s_{\mathrm{nat},c}(x) = \sum_{q=1}^{|\nu(c)|}\log p_\theta \left(\nu(c)_q \mid x,\nu(c)_{<q}
    \right).
    \label{eq:appendix-native-score}
\end{equation}

The native prediction is \(\arg\max_{c\in\mathcal{C}_d}s_{\mathrm{nat},c}(x)\). A supervised linear probe maps a role-conditioned representation to task scores:
\begin{equation}
    \mathbf{s}^{(s)}_{\mathrm{probe}}(x) = W_s\phi_{\ell,s}(x) + b_s
    \label{eq:appendix-probe}
\end{equation}

where \(W_s\) and \(b_s\) are the weight matrix and bias fitted for token role \(s\), and \(\mathbf{s}^{(s)}_{\mathrm{probe}}(x)\) holds one score per class. For binary tasks, \(s_{\mathrm{probe}}(x)\) denotes the scalar harmful-class decision score.

For the residual reconstruction intervention, the activation passed to the remaining transformer blocks is
\begin{equation}
    h'_{\ell,t} = h_{\ell,t} + \alpha_{\mathrm{hook}} \left(\widehat{h}^{\,\mathrm{joint}}_{\ell,t} -h_{\ell,t}\right),
    \qquad
    \alpha_{\mathrm{hook}}\in[0,1]
    \label{eq:appendix-reconstruction-hook}
\end{equation}

where \(h'_{\ell,t}\) is the edited state and \(\alpha_{\mathrm{hook}}\) is the reconstruction strength: \(\alpha_{\mathrm{hook}}=0\) leaves the native computation unchanged and \(\alpha_{\mathrm{hook}}=1\) fully replaces \(h_{\ell,t}\) with its joint reconstruction. The reported Qwen hook uses \(\alpha_{\mathrm{hook}}=0.35\).

Section~\ref{sec:latent_signal} introduces the direct logit router; Appendix~\ref{app:recovery-details} specifies its calibration statistics, signed task margin, mixing coefficient, and locked selection procedure.

\paragraph{Hardware and runtime.}
Experiments were run on a shared compute cluster with access to \(2\times\) NVIDIA B200 and \(2\times\) NVIDIA A100 GPUs; individual jobs used between one and four accelerators. Activation caching was the
dominant computational cost: the full Gemma/Qwen dense cache required approximately three to four 24-hour Slurm runs, followed by residual-cache construction. Most individual probe-fitting, SAE-training, inference, routing, causal-intervention, and LoRA runs completed within 12 hours, although the largest MMHS150K sweeps required up to 24 hours.

\section{Dataset and Evaluation Details}
\label{app:data-eval}

\paragraph{Primary benchmarks.}
CrisisHateMM contains text-embedded images from the Russia--Ukraine conflict annotated for hate speech and attack targets~\citep{bhandari2023crisishatemm}. FHM is a binary hate-speech benchmark constructed with benign image and text confounders intended to reduce unimodal biases~\citep{kiela2020hateful}. MAMI evaluates misogyny identification in memes~\citep{gasparini2022benchmark}; HarMeme annotates COVID-19 memes for harmfulness severity and targeted social entities~\citep{pramanick2021detecting}; MMHS150K contains paired image-text social-media posts labeled as \textsc{NotHate} or one of five hate categories~\citep{gomez2020exploring}; and MultiOFF consists of US-election memes annotated for offensiveness~\citep{suryawanshi2020multimodal}. Table~\ref{tab:primary_datasets} reports the frozen split used for each headline binary task.

\paragraph{Binary and fine-grained targets.}
The headline benchmark contains one binary task per dataset. For HarMeme, the two harmfulness categories are merged into \textsc{Harmful}, while the harmless class becomes \textsc{Not Harmful}. For MMHS150K, \{\textsc{Racist}, \textsc{Sexist}, \textsc{Homophobe}, \textsc{Religion}, \textsc{OtherHate}\} are merged into \textsc{Hate}, with \textsc{NotHate} retained as the negative class. The original three-class HarMeme task, six-class MMHS150K taxonomy, and CrisisHateMM target-attribution task are treated as secondary fine-grained evaluations and are excluded from the primary six-task scores.

\paragraph{Multilingual robustness datasets.}
EXIST-2025 contains English and Spanish memes annotated for sexism under a learning-with-disagreement framework~\citep{10.1007/978-3-032-04354-2_16}. Because official test labels were unavailable, we derive binary targets via strict majority voting on task annotations, remove tied or invalid instances, and construct a fixed, language-stratified holdout of $500$ examples, comprising $250$ English and $250$ Spanish memes, from $3,420$ valid examples. The remaining $2,920$ examples are used for probe training.

MultiBully contains Hindi-English code-mixed memes annotated for cyberbullying, sentiment, emotion, and sarcasm~\citep{10.1145/3477495.3531925}. We use its joint image-text \textsc{Bully}/\textsc{Nonbully} label rather than either unimodal annotation. After removing $53$ unreadable and $8$ missing images, a deterministic label-stratified split assigns $4,793$ of the $5,793$ valid examples to training and $1,000$ to evaluation. EXIST and MultiBully use internal holdouts and are reported only as robustness studies.

\paragraph{Model selection and locked evaluation.}
Probes and trainable adapters are fitted only on task training data. Any data-dependent choice, including layer, token role, representation, checkpoint, decision threshold, or routing coefficient, is made using training or calibration data and frozen before the reporting split is evaluated. When official validation and test sets both exist, validation is used for calibration and test for reporting. Methods requiring additional scalar calibration divide the available held-out data into disjoint calibration and reporting partitions. A result described as \emph{locked} uses no reporting example for threshold selection, checkpoint selection, or other post-hoc tuning.

\paragraph{Native decoding and label normalization.}
Genuinely binary tasks use constrained positive-versus-negative label scoring. For datasets whose original annotation space is multiclass, including HarMeme and MMHS150K, the LVLM first scores every original class label and then applies the fixed binary mapping described above. This avoids evaluating a multiclass dataset through a semantically mismatched yes/no question about one arbitrarily selected class. Primary evaluations use the original image and natively supplied OCR where available. OCR removal, image blanking, and image permutation are used only as paired robustness controls.

This native protocol deliberately measures a fixed zero-shot decision rule, not the best achievable prompted performance. Iterative rationale, few-shot, retrieval-augmented, or prompt-evolution methods such as Evolver~\citep{huang-etal-2025-evolver} may improve the emitted decision by changing the inference computation. We have not evaluated those methods here. The reported gap is a routing failure under the specified constrained-scoring protocol; it does not establish an architectural impossibility that no prompting strategy can bypass.

\paragraph{Metrics and comparison scope.}
The primary metric is macro-averaged F1, which weights each class equally despite the substantial class and prediction imbalance in several datasets. Accuracy, class-wise precision and recall, predicted
class frequencies, and paired uncertainty analyses are treated as secondary diagnostics. 

Within each model family, native, reconstruction-hook, probe, routed, and adapted predictions are compared on identical example IDs. Qwen's primary probes fix the model, layer-20 public base-SAE dictionary, and linear readout family across all six tasks, although the token role and task-specific classifier parameters vary. Gemma instead uses the calibration-selected sparse representation for each task, including single-layer, residual, and cross-layer dictionaries at different layers. Gemma and Qwen also differ in model scale, instruction-tuning status, SAE source, and dictionary construction. For these reasons, the within-task native--probe gaps are the controlled comparisons; cross-family agreement is interpreted as replication of the phenomenon, not as evidence that either model or SAE regime is superior.


\section{Additional Broad-Comparison Analysis}
\label{app:broad-comparison-details}

\paragraph{Task-level readout gaps.}
The positive sparse-readout advantage is not concentrated in one benchmark. For Gemma-3-4B-IT, the largest improvements over native macro-F1 occur on MMHS150K (\(+0.338\)), MAMI (\(+0.322\)), and CrisisHateMM (\(+0.213\)); the gains on FHM (\(+0.055\)), HarMeme (\(+0.057\)), and MultiOFF (\(+0.104\)) are smaller but remain positive. For Qwen3.5-9B-Base, the largest gains occur on CrisisHateMM (\(+0.550\)), HarMeme (\(+0.416\)), MultiOFF (\(+0.396\)), and MAMI (\(+0.348\)), with smaller positive gains on FHM (\(+0.075\)) and MMHS150K (\(+0.064\)). The unweighted mean gives every dataset equal weight, which keeps MMHS150K from dominating the aggregate despite its much larger reporting split.

\paragraph{Residual reconstruction.}
The Qwen reconstruction hook raises mean macro-F1 from \(0.432\) to \(0.486\), compared with \(0.740\) for the external SAE probe. Its aggregate recovery fraction is \[\frac{0.486-0.432}{0.740-0.432}=0.176.\] The hook improves CrisisHateMM, FHM, MAMI, MMHS150K, and MultiOFF, while remaining unchanged on HarMeme. Reconstruction can make some internally represented information behaviorally useful, but most of the probe-accessible signal is not recovered after the modified activation passes through the model's original upper layers and output head.

\paragraph{Public base SAE versus residual SAE.}
For Qwen, the public base-SAE and residual-SAE probes obtain nearly identical six-task means of \(0.740\) and \(0.737\), respectively. The public base SAE is stronger on CrisisHateMM, MultiOFF, HarMeme, and MMHS150K, while the residual SAE is stronger on FHM and MAMI. We retain the public base SAE as the primary interpretability object because it provides one standardized feature space across all six tasks, not because residualization is uniformly weaker. Their near-tie also shows that the broad readout gap is not specific to one SAE construction.

\paragraph{Fine-grained targets.}
The original three-class HarMeme and six-class MMHS150K evaluations are excluded from the primary six-task mean because they require distinguishing harmfulness severity or hate subtypes rather than the
coarser harmful-versus-benign boundary. Complete aggregate and class-wise results are reported in Appendix~\ref{app:fine-grained}.

\section{Fine-Grained Harmful-Content Classification}
\label{app:fine-grained}

\paragraph{Aggregate fine-grained results.}
Table~\ref{tab:fine_grained_results} reports the locked held-out comparison for the original HarMeme and MMHS150K taxonomies. These results are kept separate from the binary benchmark because the fine-grained tasks require distinguishing severity levels or hate subtypes rather than detecting harmful content alone.

\begin{figure}[ht]
    \centering
    \includegraphics[width=0.95\textwidth]{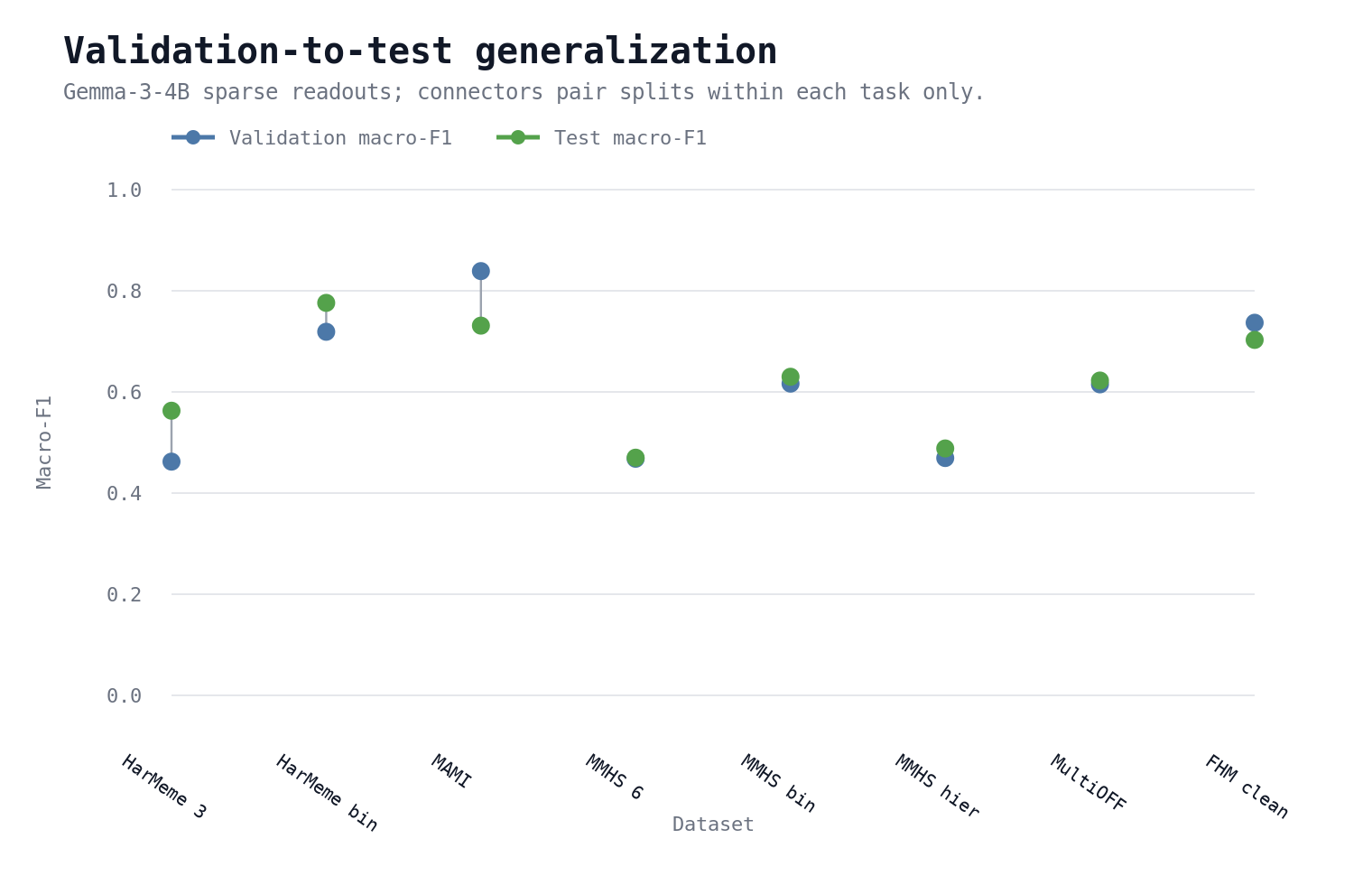}
    \caption{\textbf{Gemma validation-to-test generalization.} Paired dots preserve the original eight task-level macro-F1 comparisons; connectors join only the two splits of the same categorical task. MAMI drops from $0.8390$ to $0.7311$, whereas MMHS150K and MultiOFF transfer more closely. These are descriptive split differences, not paired-example uncertainty intervals.}
  \label{fig:val-to-test-generalization}
\end{figure}

\begin{table*}[ht]
    \centering
    \caption{Fine-grained macro-F1. For Gemma, \emph{Selected sparse} denotes the frozen validation-selected residual-SAE readout. For Qwen, the public base SAE is the primary sparse representation and the residual SAE is an ablation. The binary counterpart uses the corresponding sparse representation on the collapsed binary task. Qwen binary values are shown as base-SAE/residual-SAE.}
    \label{tab:fine_grained_results}
    \setlength{\tabcolsep}{4.5pt}
    \small
    \begin{tabular}{@{}cccccccc@{}}
        \toprule
        Family & Task & \shortstack[c]{Test\\ Count} & Native & Hook & Base SAE & \shortstack[c]{Selected/\\residual SAE} & \shortstack[c]{Binary\\ counterpart} \\
        \midrule

        \multirow{3}{*}{Gemma}
        & HarMeme 3-class & 354 & 0.3538 & 0.3103 & 0.5432 & \textbf{0.5627} & 0.7760 \\

        & \shortstack[l]{MMHS150K\\ 6-class, flat} & 10,000 & 0.3168 & 0.3118 & 0.3588 & 0.4699 & 0.6299 \\

        & \shortstack[l]{MMHS150K\\ 6-class,\\ hierarchical} & 10,000 & \textsc{NA}\textsuperscript{a} & 0.3609 & 0.4105 & \textbf{0.4882} & 0.6299 \\

        \midrule
        \multirow{2}{*}{Qwen} & HarMeme 3-class & 354 & 0.2626 & 0.2626 & 0.5292 & \textbf{0.5476}
        & 0.8096 / 0.7998 \\

        & MMHS150K 6-class & 10,000 & 0.1661 & 0.2071 & \textbf{0.3693} & 0.3602 & 0.5871 / 0.5731 \\

        \bottomrule
    \end{tabular}

    \vspace{3pt}
    \begin{minipage}{0.97\textwidth}
        \footnotesize
        \textsuperscript{a}
        The hierarchical hook/base-SAE runs use a genuine two-stage readout: `NotHate' versus `Hate', then a five-way hate-type classifier when stage 1 selects `Hate'. Native hierarchical remains `NA' because no corresponding two-stage native LVLM decoder was run.
    \end{minipage}
\end{table*}

The sparse readouts remain substantially stronger than native generation on the fine-grained tasks, but the gap between binary and fine-grained performance is large. For Gemma, the selected binary HarMeme readout reaches \(0.7760\), compared with \(0.5627\) for the selected three-class readout, a difference of \(0.2133\). These use the image and generated tokens, respectively, so the difference does not isolate label collapse in a fixed classifier. The analogous MMHS150K differences are \(0.1600\) relative to the flat six-class probe and \(0.1417\) relative to the hierarchical probe. For Qwen's primary base-SAE representation, the binary readouts exceed the fine-grained readouts by \(0.2804\) on HarMeme and \(0.2178\) on MMHS150K. The residual-SAE differences are \(0.2522\) and \(0.2129\), respectively.

\paragraph{Class-wise error structure.}
Table~\ref{tab:fine_grained_classwise} reports class-level metrics from the Gemma evaluation summary. The MMHS150K values correspond to the hierarchical six-class readout, whose class F1 values average to its reported macro-F1 of \(0.4882\). The HarMeme values correspond to the selected generated-token three-class readout.

\begin{table}[!ht]
    \centering
    \caption{Class-level metrics explicitly recorded for the locked Gemma fine-grained evaluation.}
    \label{tab:fine_grained_classwise}
    \small
    \begin{tabular}{@{}cccccc@{}}
        \toprule
        Task & Class & Precision & Recall & F1 & Support / diagnostic \\
        \midrule

        \multirow{3}{*}{\shortstack[c]{HarMeme\\ 3-class}} & not harmful & 0.9337 & 0.7957 & 0.8592 & 230 gold; 183 TP \\

        & somewhat harmful & 0.5986 & 0.8544 & 0.7040 & 103 gold; 88 TP \\
        & very harmful & 0.1818 & 0.0952 & 0.1250 & 21 gold; 2 TP \\

        \midrule
        \multirow{6}{*}{\shortstack[c]{MMHS\\ hierarchical}} & Homophobe & 0.6202 & 0.6763 & \textbf{0.6470} & 1,007 gold; 681 TP \\

        & NotHate & 0.7275 & 0.7192 & \textbf{0.7233} & 6,090 gold; 4,380 TP \\
        & OtherHate & 0.4446 & 0.5356 & 0.4858 & 801 gold; 429 TP \\
        & Racist & 0.4458 & 0.3594 & 0.3979 & 1,614 gold; 580 TP \\
        & Religion & 0.4444 & 0.3333 & 0.3810 & \shortstack[c]{24 gold; 8 TP;\\ low-frequency class} \\
        & Sexist & 0.2613 & 0.3362 & 0.2941 & 464 gold; 156 TP \\

        \bottomrule
    \end{tabular}
\end{table}

HarMeme's aggregate accuracy of \(0.7712\) masks a severe minority-class failure. Only \(2\) of the \(21\) \texttt{very harmful} examples are classified correctly, and the class F1 is \(0.125\). The model captures the broad harmfulness boundary much more reliably than the distinction between \texttt{somewhat harmful} and \texttt{very harmful}.

MMHS150K shows a less concentrated but still uneven error profile. \texttt{NotHate} and \texttt{Homophobe} are the strongest classes, with F1 values of \(0.723\) and \(0.647\). \texttt{Sexist} is the weakest at \(0.294\), while \texttt{Racist}, \texttt{Religion}, and \texttt{OtherHate} remain between \(0.381\) and \(0.486\). The hierarchical formulation improves aggregate macro-F1 from \(0.4699\)
to \(0.4882\), but it does not eliminate the substantial variation between hate subtypes.

\section{Token-Role and Representation Ablations}
\label{app:role-representation-ablation}

\paragraph{Calibration-only role selection.}
For each task, candidate probes over prompt/OCR, image, generated, pre-generation, and all-token representations use the same training examples and probe family. 


\begin{table}[t]
    \centering
    \caption{\textbf{Gemma-3-4B token-role ablation on the validation split.} Values are macro-F1 used for representation selection, not final reporting. Positive \(\Delta_{\mathrm{gen-img}}\) indicates that generated-token states outperform image-position states.}
    \label{tab:gemma-token-role-ablation}
    \small
    \setlength{\tabcolsep}{5pt}
    \renewcommand{\arraystretch}{0.96}
    \begin{tabular}{@{}lrrr@{}}
        \toprule
        Dataset & Image & Generated & \(\Delta_{\mathrm{gen-img}}\) \\
        \midrule
        CrisisHateMM & 0.817 & 0.811 & \(-0.006\) \\
        FHM          & 0.645 & 0.651 & \(+0.006\) \\
        MAMI         & 0.789 & 0.839 & \(+0.050\) \\
        HarMeme      & 0.719 & 0.668 & \(-0.051\) \\
        MMHS150K     & 0.531 & 0.617 & \(+0.086\) \\
        MultiOFF     & 0.600 & 0.615 & \(+0.016\) \\
        \bottomrule
    \end{tabular}
\end{table}

\paragraph{Task-dependent token roles.}
Table~\ref{tab:gemma-token-role-ablation} rules out a simple account in which harmful content information is consistently concentrated in either image-origin or generated states. Image features are strongest for HarMeme and marginally stronger for CrisisHateMM. Generated features outperform image features on MAMI, MMHS150K, and MultiOFF, while the two FHM readouts differ by less than one macro-F1 point.

The clearest shift occurs on MMHS150K. Moving from image to generated states raises binary macro-F1 from \(0.531\) to \(0.617\). On the original six-class taxonomy, the corresponding increase is from \(0.232\) to \(0.468\). The decoding trajectory makes hate subtypes such as \textsc{Racist}, \textsc{Homophobe}, and \textsc{NotHate} considerably easier to separate than they are in image-position states. This result does not imply that generated tokens supply the gold label; rather, the model's own classification or explanation trajectory reorganizes the input into a more label-separable representation.

MAMI exhibits a different pattern. Its prompt/OCR probe reaches \(0.845\) macro-F1, slightly exceeding the generated-token result and clearly exceeding the image-position result of \(0.789\). The strongest calibration signal comes from text-bearing states. Prompt positions may incorporate visual information through contextual processing, so this is not a text-only intervention, nor is it direct evidence of visual misogyny recognition. The subsequent decrease from validation to test (seen in Figure~\ref{fig:val-to-test-generalization}) indicates limited generalization but does not, on its own, identify reliance on lexical cues.

\paragraph{Pre-generation and post-generation interpretations.}
Generated-token representations are observed only after answer generation begins and may encode an emerging label, explanation, or confidence. We use them to locate when the representation becomes separable, not as sole evidence that the same information was available before decoding.

Pre-generation controls show that the broad readout gap does not come from probing the model's own answer. Image-position probes are selected for CrisisHateMM and binary HarMeme and retain substantial advantages over native output on their locked reporting splits. The selected Gemma FHM readout uses prompt and image positions from a cross-layer representation rather than generated states. Both pre-generation and post-generation states contain useful harmful content information, although the strongest role varies by task.

\paragraph{FHM multi-role controls.}
FHM was designed with benign image and text confounders, so a useful classifier must distinguish examples whose labels change when one modality is held constant~\citep{kiela2020hateful}. Gemma's image-generated and all-token layer-22 probes remain in a similar performance range. Its final 4B candidate instead combines prompt--image crosscoder features with complementary layer-22 and layer-29 readouts.

Qwen provides a separate control using its public layer-20 base SAE. Image and prompt-position probes obtain \(0.628\) and \(0.631\) macro-F1, respectively, while their concatenation reaches \(0.641\). A rank-16 bilinear probe falls to \(0.528\). Both roles carry useful additive information, but this pooled bilinear parameterization does not recover a useful interaction. More generally, using multiple modality-associated representations does not, by itself, establish a genuine cross-modal interaction~\citep{hessel-lee-2020-multimodal}. Section~\ref{subsec:fhm_bilinear} tests this question using matched confounders and a stronger Gemma-3-12B representation.

\paragraph{Sparse versus dense representations.}
We compare each sparse readout with a probe over the corresponding role-pooled dense hidden state. For Gemma, sparse decomposition improves macro-F1 by \(0.283\) on CrisisHateMM, \(0.041\) on six-class
MMHS150K, and \(0.022\) on FHM. Its effect on MAMI, HarMeme, and MultiOFF ranges from \(-0.006\) to \(+0.008\). Random sparse controls remain near chance, ruling out dimensionality or sparsity alone as the
source of the strong probe results.

These controls indicate that most harmful content information is already present in the dense residual stream. Sparse representation can make that information easier to separate on selected tasks, but its
primary value for the subsequent analysis is that it expresses the signal in a coordinate system supporting feature-level output alignment, ablation, and activation patching. The complementary Qwen
public-base versus residual-SAE comparison is reported in Appendix~\ref{app:broad-comparison-details}.

\section{Feature Alignment and Causal Intervention Details}
\label{app:causal-details}

Silent knockout tests whether the discriminative set is necessary for the probe, whereas routed patching tests whether an independently selected feature set can control the monitored native output. Because the patch broadcasts max-pooled donor values across target positions, it does not reproduce a naturally occurring activation pattern. We interpret only the relative magnitudes of the paired score changes, not their signs or whether the intervention improves correctness.

\paragraph{Static feature-output alignment.}
For an SAE feature with decoder direction \(d_j\in\mathbb{R}^{d_\ell}\), we measure its direct alignment with the yes/no output channel. Let
\begin{equation}
    m_{\mathrm{yn}}(x) = \ell_{\mathrm{yes}}(x)-\ell_{\mathrm{no}}(x)
    \label{eq:appendix-yes-no-margin}
\end{equation}

denote the monitored output margin, where \(\ell_v(x)\) is the final-position logit for vocabulary token \(v\) (not a layer index). With the final normalization scale folded into the unembedding \(W_U^f\), and with \(e_v\) the one-hot vocabulary vector for token \(v\), the corresponding hidden-space direction and feature effect are
\begin{equation}
    u_{\mathrm{yn}} = W_U^{f\top} \left(e_{\mathrm{yes}}-e_{\mathrm{no}}\right),
    \qquad
    a_j = d_j^\top u_{\mathrm{yn}}.
    \label{eq:appendix-output-alignment}
\end{equation}

The signed agreement between this effect and task-\(d\) probe weight \(w_{d,j}\) is
\begin{equation}
    q_{d,j} = \operatorname{sign}(w_{d,j})a_j.
    \label{eq:appendix-signed-alignment}
\end{equation}

A probe-important feature is called \emph{aligned} when \(q_{d,j}\geq\tau\), \emph{misaligned} when
\(q_{d,j}\leq-\tau\), and \emph{output-silent} when \(|a_j|<\tau\), using \(\tau=0.10\). Silence is defined only relative to this output anchor; it does not imply that the feature is globally
inactive or incapable of affecting another output channel.

For Qwen's public base SAE, the aligned subset is empty for CrisisHateMM, FHM, MultiOFF, HarMeme, and MMHS150K, and contains only \(1\%\) of analyzed feature records for MAMI. No task has a substantial misaligned subset. The secondary residual SAE gives the same qualitative result. Because Equations~\ref{eq:appendix-yes-no-margin} and~\ref{eq:appendix-output-alignment} bypass the intervening transformer blocks, we use them only to select features, not as complete causal attributions~\citep{elhage2021mathematical,belrose2023eliciting}.

\paragraph{Independent feature-set construction.}
The routed set is selected independently of labels and probe weights:
\begin{equation}
    \mathcal{R}_{\mathrm{yn}} = \operatorname{TopK}_k\left(\left\{|d_j^\top u_{\mathrm{yn}}|\right\}_{j=1}^{m}\right).
    \label{eq:appendix-routed-set}
\end{equation}

For task \(d\), the silent discriminative set is
\begin{equation}
    \mathcal{S}_d = \left\{j\in \operatorname{TopK}_k\left(\{|w_{d,j}|\}_{j=1}^{m}\right):|a_j|<\tau
    \right\}.
    \label{eq:appendix-silent-set}
\end{equation}

All reported Qwen interventions use \(k=20\), and \(m\) is the SAE dictionary width (\(65{,}536\) for the Qwen base SAE). Thus, \(\mathcal{S}_d\) is selected for task discrimination subject to weak output alignment, while \(\mathcal{R}_{\mathrm{yn}}\) is selected for output alignment without using the task labels.

\paragraph{Paired intervention protocol.}
For silent-feature knockout, we select up to \(N_{\mathrm{KO}}=15\) reporting examples per task whose gold label is the positive/harmful class, for which the frozen probe is correct, and the native model is wrong. For each example, we set the \(k_{\mathrm{causal}}=20\) coordinates in \(\mathcal{S}_d\) to zero at every token position in the layer-20 SAE code. We then recompute the role-pooled probe score, propagate the edited representation through the remaining transformer blocks, and recompute the monitored native margin. The original and intervened scores are paired measurements on the same example.

For routed-feature patching, we use up to \(N_{\mathrm{pair}}=5\) eligible donor--target pairs per task. The donor \(x'\) supplies the max-over-token activation \(\max_{t'}z_{t',j}(x')\) for each of the $20$ features \(j\in\mathcal{R}_{\mathrm{yn}}\). This value overwrites feature \(j\) at every token position of the target \(x\), after which the target forward pass continues through the remaining model. The target is the experimental unit whose probe score and native margin are measured; the donor supplies only the replacement feature values. 


\paragraph{Knockout and donor patching.}
Following causal-mediation and activation-patching analyses~\citep{vig2020investigating,meng2022locating,marks2025sparse}, the knockout intervention removes a selected set \(S\) from the layer-20 SAE code \(z_{t,j}(x)\) of feature \(j\) at position \(t\), where we drop the fixed layer index:
\begin{equation}
    z^{-S}_{t,j}(x) =
    \begin{cases}
        0, & j\in S,\\
        z_{t,j}(x), & j\notin S.
    \end{cases}
    \label{eq:appendix-feature-knockout}
\end{equation}

For donor \(x'\) and target \(x\), routed-feature patching assigns the donor's max-pooled activation to the selected target features:
\begin{equation}
    z^{\,x\leftarrow x'}_{t,j} =
    \begin{cases}
        \displaystyle\max_{t'} z_{t',j}(x'),
            & j\in\mathcal{R}_{\mathrm{yn}},\\
        z_{t,j}(x),
            & j\notin\mathcal{R}_{\mathrm{yn}}.
    \end{cases}
    \label{eq:appendix-feature-patching}
\end{equation}

The patched value is broadcast over target positions. This is useful for testing whether the selected feature family can control a downstream quantity, but it does not reproduce the donor's natural
position-dependent activation pattern.

For intervened computation \(\widetilde{x}\), we record
\begin{equation}
    \Delta_{\mathrm{probe}} = s_{\mathrm{probe}}(\widetilde{x}) - s_{\mathrm{probe}}(x),
    \qquad
    \Delta_{\mathrm{native}} = m_{\mathrm{yn}}(\widetilde{x}) - m_{\mathrm{yn}}(x)
    \label{eq:appendix-intervention-effects}
\end{equation}

and summarize the mirror-image interventions using
\begin{equation}
    \rho_{\mathcal{S}} = \frac{\mathbb{E}|\Delta_{\mathrm{probe}}|}{\mathbb{E}|\Delta_{\mathrm{native}}|
    },
    \qquad
    \rho_{\mathcal{R}} = \frac{\mathbb{E}|\Delta_{\mathrm{native}}|}{\mathbb{E}|\Delta_{\mathrm{probe}}|
    }.
    \label{eq:appendix-sensitivity-ratios}
\end{equation}

Here \(s_{\mathrm{probe}}\) is the scalar harmful-class probe score recomputed from the edited code, \(\mathbb{E}\) averages over the eligible examples or donor--target pairs, \(\rho_{\mathcal{S}}\) is reported for silent-feature knockout, and \(\rho_{\mathcal{R}}\) for routed-feature patching.

\paragraph{Qwen intervention results.}
Silent-feature knockout uses at most 15 eligible examples per task. Where eligible examples exist, removing only 20 features changes the probe score by approximately \(2.0\)--\(2.5\) units while changing the monitored yes/no margin by only \(0.04\)--\(0.08\). This gives probe-to-margin ratios of \(24.2{:}1\) to \(62.8{:}1\). For FHM, the monitored margin is the actual native decision channel. For HarMeme and MMHS150K, it is only a diagnostic because their native predictions use full-label scoring.

We bootstrap unique examples rather than feature edits. For FHM knockout, the signed mean changes and 95\% intervals are \(\Delta_{\mathrm{probe}}=-2.512\;[-2.769,-2.250]\) and \(\Delta_{\mathrm{native}}=-0.033\;[-0.063,-0.008]\); for HarMeme they are \(-2.500\;[-2.621,-2.385]\) and \(-0.027\;[-0.049,-0.006]\). These intervals support pathway separation for the larger knockout cohorts but do not remove the scale dependence of \(\rho_{\mathcal S}\) or \(\rho_{\mathcal R}\).

Routed-feature patching uses at most five eligible donor-target pairs. On the four literal yes/no tasks, it changes the native margin by \(0.34\)--\(1.09\) while changing the probe score by only
\(0.006\)--\(0.062\), yielding native-to-probe ratios from \(16.2{:}1\) to \(140.2{:}1\). The five-pair estimates are exploratory: for FHM, routed-patch \(\Delta_{\mathrm{native}}=+0.338\) has a 95\% interval of \([+0.175,+0.438]\), while the corresponding small cohorts on other tasks remain too imprecise for broad population claims. The MMHS150K routed result is excluded from the headline comparison because both its output channel and routed feature set are mismatched to the task's label-score decision rule. These effects establish control of the monitored output, not an improvement in prediction correctness.

\paragraph{Why silent-feature donor patching is not conclusive evidence.}
Broadcasting a donor's max-pooled feature value to every target token position can create activation patterns that do not occur during ordinary inference. This intervention produces inconsistent signs and, for CrisisHateMM, an unusually large movement of the monitored margin. We therefore use silent-feature knockout to establish probe dependence and routed-feature patching to establish output control. Silent-feature patching is retained in Table~\ref{tab:full_causal_interventions} as a transparency and sanity-check result, not as primary evidence.

\paragraph{Gemma replication and limits.}
For Gemma's layer-22 residual-SAE dictionary, the top probe-derived and directly routed sets have zero overlap on all four analyzed tasks. Perturbing the silent set changes the probe score \(13\)--\(180\) times more than perturbing the routed set. We additionally compute an upper-layer-aware output-sensitivity diagnostic:
\begin{equation}
    a^{J}_{d,j} = d_j^\top \mathbb{E}_{x}\left[\nabla_{h_\ell}m_d(x)\right].
    \label{eq:appendix-jacobian-alignment}
\end{equation}

where \(\nabla_{h_\ell}m_d(x)\) is the gradient of the signed native task margin \(m_d\) with respect to the layer-\(\ell\) residual state, averaged over the analyzed examples. Unlike \(a_j\), this quantity includes the effect of the upper transformer blocks to first order.

This Jacobian-based~\citep{gurnee2026verbalizable} analysis also places the strongest probe directions outside the dominant output-routing set. Native-output mediation is less uniform: routed-feature ablation has a larger effect than silent-feature ablation on MMHS150K, but not consistently on MAMI, HarMeme, or CrisisHateMM. In short, Gemma replicates the geometric and probe-side separation, but does not support a universal two-subspace causal decomposition.



\begin{table*}[ht]
    \centering
    \caption{\textbf{Complete causal interventions for Qwen's public layer-20 base SAE.} \(\mathbb{E}|\Delta_{\mathrm{probe}}|\) and \(\mathbb{E}|\Delta_{\mathrm{native}}|\) are mean absolute changes in the external probe score and monitored yes/no margin. The final column is always written as probe sensitivity: native-margin sensitivity. Thus, \(50.2{:}1\) is probe-dominant, whereas \(1{:}16.2\) is output-dominant. YN denotes literal yes/no decoding, and LS denotes full-label scoring. Count represents the eligible examples for knockout and donor-target pairs for patching.}
    \label{tab:full_causal_interventions}
    \small
    \begin{tabular}{@{}cccccc@{}}
        \toprule
        Task
        & Decode & Count & \(\mathbb{E}|\Delta s_{\mathrm{probe}}|\) & \(\mathbb{E}|\Delta m_{\mathrm{yn}}|\) & Probe:YN \\
        \midrule

        \multicolumn{6}{@{}l}{
            \textit{A. Silent-feature knockout}
        }\\[0pt]

        CrisisHateMM & YN & --\textsuperscript{a} & -- & -- & -- \\

        FHM & YN & 15 & 2.5121 & 0.0500 & \textbf{50.2:1} \\
        MAMI & YN & --\textsuperscript{a} & -- & -- & -- \\
        HarMeme & LS\textsuperscript{\(\dagger\)} & 15 & 2.4995 & 0.0398 & \textbf{62.8:1} \\
        MMHS150K & LS\textsuperscript{\(\dagger\)} & 15 & 2.0190 & 0.0833 & \textbf{24.2:1} \\
        MultiOFF & YN & --\textsuperscript{a} & -- & -- & -- \\

        \addlinespace[3pt]
        \multicolumn{6}{@{}l}{
            \textit{B. Silent-feature donor patch}
        }\\[0pt]

        CrisisHateMM & YN & 5 & 0.3416 & 1.5377 & 0.2:1 \\
        FHM & YN  & 5 & 0.4010 & 0.1373 & 2.9:1 \\
        MAMI & YN & 5 & 0.6764 & 0.1124 & 6.0:1 \\
        HarMeme & LS & --\textsuperscript{b} & -- & -- & -- \\
        MMHS150K & LS\textsuperscript{\(\dagger\)} & 5 & 0.4248 & 0.2737 & 1.6:1 \\
        MultiOFF & YN & 5 & 0.5354 & 0.1253 & 4.3:1 \\

        \addlinespace[3pt]
        \multicolumn{6}{@{}l}{
            \textit{C. Routed-feature donor patch}
        }\\[0pt]

        CrisisHateMM & YN & 5 & 0.0125 & 0.3998 & \textbf{1:32.0} \\
        FHM & YN & 5 & 0.0208 & 0.3376 & \textbf{1:16.2} \\
        MAMI & YN & 5 & 0.0624 & 1.0880 & \textbf{1:17.4} \\
        HarMeme & LS & --\textsuperscript{b} & -- & -- & -- \\
        MMHS150K & LS\textsuperscript{\(\ddagger\)} & 5 & 0.3191 & 0.0900 & 3.5:1 \\
        MultiOFF & YN & 5 & 0.0065 & 0.9116 & \textbf{1:140.2} \\

        \bottomrule
    \end{tabular}

    \vspace{3pt}
    \begin{minipage}{0.97\textwidth}
        \footnotesize
        \textsuperscript{a}
        No eligible gold-positive, probe-correct, native-wrong examples occurred under the Intervention~A selection rule. The native classifiers for these tasks already predicted the
        positive class at a high rate, leaving few or no relevant false negatives.

        \textsuperscript{b}
        No eligible patch targets occurred for HarMeme because the native Qwen classifier predicted \texttt{not harmful} for every reporting example.

        \textsuperscript{\(\dagger\)}
        HarMeme and MMHS150K use label-score decoding rather than literal yes/no decoding. Their reported
        \(\Delta m_{\mathrm{yn}}\) therefore measures a diagnostic yes/no direction, not the complete native task decision.

        \textsuperscript{\(\ddagger\)}
        In addition to the diagnostic-margin caveat, the routed feature set itself was selected using the yes/no unembedding direction. It is therefore mismatched to MMHS150K's label-score channel and is not included in the routed-feature headline conclusion.
    \end{minipage}
\end{table*}

\section{FHM Low-Rank Cross-Modal Readout Details}
\label{app:fhm-low-rank-details}

\paragraph{Representation construction.}
We use the Gemma-3-12B layer-31 sparse representation. The gated backbone receives the train-only MaxAbs-scaled all-token vector \(x_{\mathrm{all}}\in\mathbb{R}^{65{,}536}\). A train-only confounder audit separately selects \(K=128\) features for the interaction term, after which image-position and prompt-position activations are feature-wise max-pooled and MaxAbs-scaled:
\[x_{\mathrm{img}},x_{\mathrm{prompt}}\in\mathbb{R}^{K}.\]

The prompt representation includes both the task instruction and provided OCR. Image and prompt refer to token origin; by layer 31, both states may contain information exchanged through the transformer.

\paragraph{Gated pairwise baseline.}
Two one-hidden-layer ReLU MLP heads, each with hidden width 128, are trained on the same all-token vector but receive their respective same-image or same-text pair-ranking losses:
\begin{equation}
    s_{\mathrm{img-pair}}(x) = f_{\mathrm{img-pair}}(x_{\mathrm{all}}),
    \qquad
    s_{\mathrm{text-pair}}(x) = f_{\mathrm{text-pair}}(x_{\mathrm{all}}).
    \label{eq:app-fhm-role-scores}
\end{equation}

An input-dependent two-layer gate with hidden width 32 combines them:
\begin{equation}
    \begin{aligned}
    g(x) &= \sigma\!\left(f_g(x_{\mathrm{all}})\right),
    \\
    s_{\mathrm{gate}}(x) &= g(x)s_{\mathrm{img-pair}}(x) + \bigl(1-g(x)\bigr)s_{\mathrm{text-pair}}(x).
    \end{aligned}
    \label{eq:app-fhm-gated-baseline}
\end{equation}

The head names identify the pair family supplying each ranking term, not a modality-specific input. Here \(f_{\mathrm{img-pair}}\), \(f_{\mathrm{text-pair}}\), and \(f_g\) are the two MLP heads and the gate network, \(\sigma\) is the logistic sigmoid, and \(g(x)\in(0,1)\) is the weight given to the same-image head. We refer to this as a gated pairwise baseline rather than a linear model.

\paragraph{Low-rank interaction.}
The full readout is defined in Equation~\ref{eq:fhm-low-rank}. Equivalently, its interaction matrix is
\[W_{\mathrm{bil}}=UV^\top, \qquad U,V\in\mathbb{R}^{K\times r}.\]

The rank \(r\) bounds the number of multiplicative image-by-prompt directions and reduces the interaction parameter count from \(K^2\) to \(2Kr\).

\paragraph{Matched-pair objective.}
Let \(\mathcal{P}_{\mathrm{img}}\) contain opposite-label pairs sharing the same image, and let \(\mathcal{P}_{\mathrm{text}}\) contain opposite-label pairs sharing the same meme text. For \((x^+,x^-)\), \(x^+\) is the hateful member and \(x^-\) is the benign member. Training minimizes
\begin{equation}
    \mathcal{L} = \mathcal{L}_{\mathrm{cls}} + \sum_{c\in\{\mathrm{img},\mathrm{text}\}} \frac{\lambda_c}{|\mathcal{P}_c|} \sum_{(x^+,x^-)\in\mathcal{P}_c}
    \left[\delta-s_r(x^+)+s_r(x^-)\right]_+
    \label{eq:app-fhm-pair-loss}
\end{equation}

where \(\mathcal{L}_{\mathrm{cls}}\) is the sample-level classification loss on \(s_r\), \(c\) indexes the two pair families, \([a]_+=\max(0,a)\), \(\delta\) is the required pair margin, and \(\lambda_c\) controls the contribution of each pair family.

\paragraph{Development and locked-test protocol.}
Model parameters are fitted on the FHM training set. Decision thresholds are selected on a fixed 94-example calibration partition and evaluated on a disjoint 375-example validation partition. Training seeds change only parameter initialization. Rank $32$ is chosen from the development rank sweep based on its accuracy-stability trade-off. Within that rank, the seed-0 checkpoint and threshold \(0.52\) are selected from calibration results and then frozen before evaluating the $1000$ test-seen examples with paired predictions.

\begin{table}[t]
    \centering
    \caption{\textbf{Gemma-3-12B FHM rank sweep.} Validation macro-F1 is measured on the same 375-example partition after threshold selection on the disjoint 94-example calibration partition. Rank-8 and rank-32 values are the mean \(\pm\) standard deviation over five seeds. Pair BC reports the mean both-correct rate for same-image and same-text opposite-label pairs. The \(\dagger\) test result is from the calibration-selected rank-32, seed-0 checkpoint, not an average over test runs.}
    \label{tab:fhm-rank-sweep}
    \small
    \setlength{\tabcolsep}{4.5pt}
    \renewcommand{\arraystretch}{0.96}
    \begin{tabular}{@{}cccc@{}}
        \toprule
        Readout
        & Validation F1 & Pair BC, Img/Txt & Test-seen F1 \\
        \midrule
        Native 12B label score & -- & -- & 0.6853 \\
        Bilinear \(r=8\) & \(0.7110\pm0.0352\) & 0.471 / 0.407 & -- \\
        Bilinear \(r=32\) & \(0.7488\pm0.0120\) & 0.555 / 0.464 & \(0.7560^\dagger\) \\
        Bilinear \(r=128\) & 0.7621 & 0.583 / 0.506 & -- \\
        \bottomrule
    \end{tabular}
\end{table}

\paragraph{Locked test comparison.}
The selected rank-32 checkpoint transfers from \(0.7591\) validation macro-F1 to \(0.7560\) on the $1000$-example paired test-seen subset. Native Gemma-3-12B obtains \(0.6853\), giving \[\Delta_{\mathrm{test}} = 0.7560-0.6853 = 0.0707.\]

A label-stratified paired bootstrap gives a 95\% confidence interval of \([0.0461,0.0954]\). An exact McNemar test favors the pair-aware bilinear readout on 103 discordant examples, compared with 37 examples favoring the native model (\(p=2.18\times10^{-8}\)). These statistics compare the complete
pair-aware readout with native generation. They should not be interpreted as the isolated effect of adding the bilinear component to the gated baseline.

\begin{figure}[ht]
  \centering
  \includegraphics[width=0.95\linewidth]{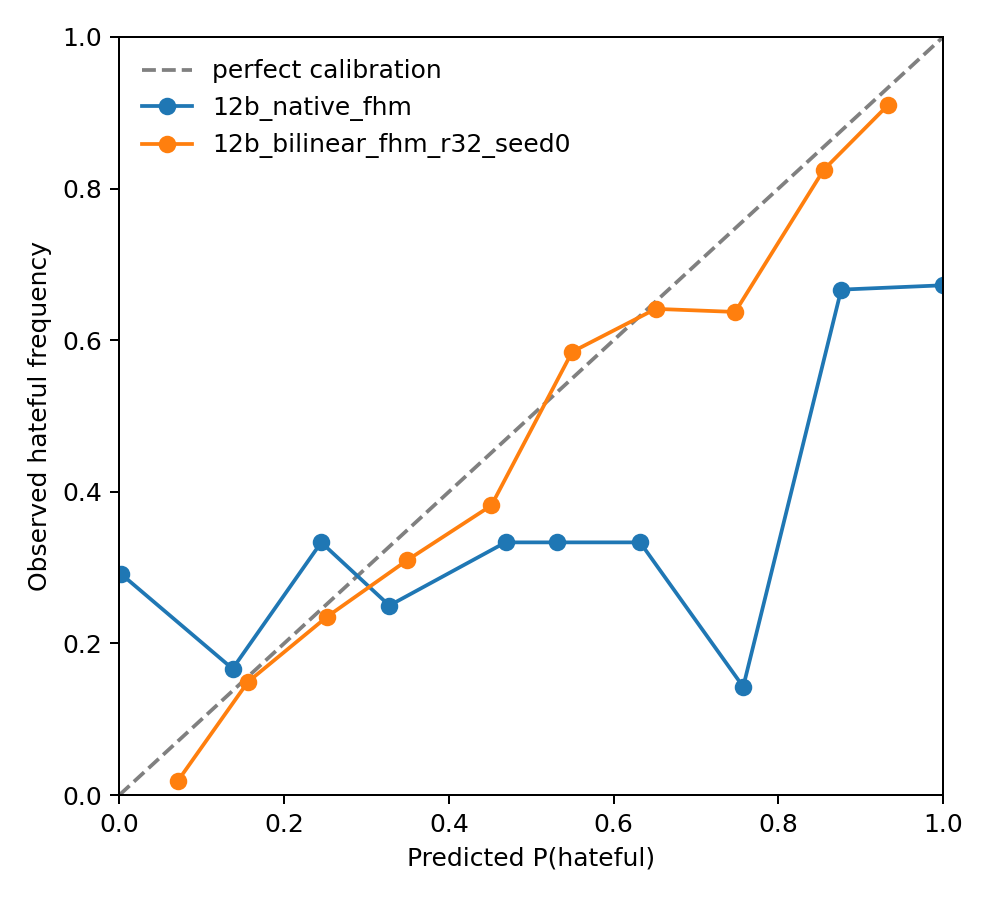}
  \caption{FHM calibration curve (Gemma-3-12B).}
  \label{fig:fhm-calibration-curve}
\end{figure}


\paragraph{Matched-confounder behavior.}
For the selected rank-32 seed, the hateful member receives the larger score in \(84.5\%\) of same-image pairs and \(84.7\%\) of same-text pairs. The stricter both-correct rates are \(58.3\%\) and \(49.4\%\), respectively. The model captures a strong relative-ordering signal when either the image or text is held fixed, while exact classification of both members remains more difficult.

\paragraph{Cross-covariance interpretation.}
For centered role vectors and binary label \(y\) (\(1=\) hateful), define the class-conditional cross-covariance difference
\begin{equation}
    \Delta\Sigma_{\mathrm{ip}} = \mathbb{E}\left[x_{\mathrm{img}}x_{\mathrm{prompt}}^\top\mid y=1\right]
    -
    \mathbb{E}\left[x_{\mathrm{img}}x_{\mathrm{prompt}}^\top\mid y=0\right].
    \label{eq:app-fhm-cross-covariance}
\end{equation}

The class separation contributed by the explicit interaction component, that is, the difference between its mean value on hateful and benign examples, is
\begin{equation}
    \Delta_{\mathrm{int}} = \gamma\operatorname{tr}\left(W_{\mathrm{bil}}^\top\Delta\Sigma_{\mathrm{ip}}
    \right).
    \label{eq:app-fhm-interaction-separation}
\end{equation}

The bilinear term thus responds to class-dependent co-variation between the two role representations, which a readout of their separate marginal activations cannot capture. Here \(\operatorname{tr}(\cdot)\) is the matrix trace and \(\gamma\) is the interaction weight from Equation~\ref{eq:fhm-low-rank}.

\paragraph{Nonlinear and factor-level controls.}
A within-role quadratic control adds squared image, prompt, and generated-feature activations without multiplying features across roles. It changes validation macro-F1 by only \(0.004\) and leaves the matched-pair both-correct rates unchanged. So generic nonlinear rescaling within an individual role does not explain the improvement.

Leave-one-rank-out analysis finds that 11 of the 32 interaction factors make a measurable contribution to validation macro-F1. Removing any single factor changes macro-F1 by at most \(0.007\). Follow-up patching and polarity tests do not support a collection of narrowly bound rules of the form ``specific image feature \(i\) plus specific text feature \(j\).'' The more conservative interpretation is a redundant, distributed interaction over broader image- and prompt-side subspaces.

\paragraph{Cross-model negative controls.}
The result does not transfer uniformly to other representations. For Gemma-3-4B, the corresponding bilinear readout obtains \(0.654\) macro-F1, below its simpler baseline at \(0.688\). For Qwen's public
layer-20 base SAE, a rank-16 bilinear readout obtains \(0.528\), below image-prompt concatenation at \(0.641\). These results limit the interaction claim to the analyzed Gemma-3-12B layer-31 representation
and pair-aware training procedure; they do not show that other models lack cross-modal structure at different layers or under different readouts.

\section{Direct-Routing and Probe-Distillation Details}
\label{app:recovery-details}

\subsection{Calibration-Only Direct Logit Routing}
\label{app:direct-routing}

For task \(d\), let \(\mathcal{D}_{d,\mathrm{cal}}\) be the calibration partition, \(m_d(x)\) the signed native decision margin, and \(u_d(x)\) the decision-function score of the frozen probe. The calibration statistics are
\begin{equation}
    \begin{aligned}
    \mu_d &= \frac{1}{|\mathcal{D}_{d,\mathrm{cal}}|} \sum_{x\in\mathcal{D}_{d,\mathrm{cal}}}u_d(x),
    \\
    \sigma_d^2 &= \frac{1}{|\mathcal{D}_{d,\mathrm{cal}}|} \sum_{x\in\mathcal{D}_{d,\mathrm{cal}}}
    \bigl(u_d(x)-\mu_d\bigr)^2,
    \\
    \widetilde{u}_d(x) &= \frac{u_d(x)-\mu_d}{\sigma_d+\epsilon}.
    \end{aligned}
    \label{eq:app-router-normalization}
\end{equation}

To isolate threshold error before introducing the probe, we select a native-only threshold on the same calibration partition,
\begin{equation}
    t_d^\star = \arg\max_t \operatorname{MacroF1}\left(
    \left\{\mathbb{I}[m_d(x)>t]\right\}_{x\in\mathcal{D}_{d,\mathrm{cal}}},
    \mathbf{y}_{d,\mathrm{cal}}\right),
    \label{eq:app-native-threshold}
\end{equation}
and freeze \(t_d^\star\) before reporting. Native AUROC evaluates ranking independently of this operating point.

The routed margin is given by
Equation~\ref{eq:direct-logit-routing}. Its mixing coefficient is selected only on the calibration partition:
\begin{equation}
    \beta_d^\star = \arg\max_{\beta\in\mathcal{A}}\operatorname{MacroF1}\left(\left\{\mathbb{I}\left[
    m_d(x)+\beta\widetilde{u}_d(x)\geq 0\right]\right\}_{x\in\mathcal{D}_{d,\mathrm{cal}}},
    \mathbf{y}_{d,\mathrm{cal}}\right),
    \label{eq:app-router-selection}
\end{equation}

where \(\mathcal{A}\) is the finite coefficient grid, \(\mathbb{I}[\cdot]\) is the indicator function, and \(\mathbf{y}_{d,\mathrm{cal}}\) are the calibration gold labels. Here \(\epsilon\) is a small numerical-stability constant and \(\widetilde{u}_d\) is the standardized probe score. The normalization statistics, coefficients, and zero decision threshold are then frozen before evaluation on the disjoint reporting partition.

\begin{table*}[ht]
    \centering
    \caption{\textbf{Native calibration and Qwen direct logit routing on disjoint reporting partitions.} A native threshold and routing coefficient are selected on the same label-stratified 30\% calibration partition and frozen on the remaining 70\%. F1 denotes macro-F1; AUC denotes AUROC. Majority predicts the calibration-set majority class. MultiOFF has complete native-score results, but no matched continuous probe-score artifact from the final cache-backed run; its unavailable cells and the five-task matched mean are marked accordingly.}
    \label{tab:direct-routing-full}
    \scriptsize
    \renewcommand{\arraystretch}{0.96}
    \begin{tabular}{@{}lcccccccc@{}}
        \toprule
        Task & Majority F1 & Native F1 & \(t_d^\star\) & Native-cal. F1 & Native AUC & Probe F1 & Probe AUC & Routed F1 \\
        \midrule
        CrisisHateMM & 0.3542 & 0.3697 &  0.7501 & 0.6189 & 0.6711 & 0.8520 & 0.9368 & 0.8417 \\
        FHM          & 0.3411 & 0.6505 &  0.9375 & 0.7094 & 0.7646 & 0.7068 & 0.8096 & 0.7403 \\
        MAMI         & 0.3333 & 0.4045 &  3.0621 & 0.7596 & 0.8426 & 0.7549 & 0.8919 & 0.7853 \\
        HarMeme      & 0.3936 & 0.3936 & -3.1078 & 0.6503 & 0.7337 & 0.8246 & 0.9074 & 0.8241 \\
        MMHS150K     & 0.3785 & 0.5016 &  0.7089 & 0.5581 & 0.5788 & 0.5807 & 0.6184 & 0.5769 \\
        MultiOFF     & 0.3772 & 0.3175 &  1.3751 & 0.5562 & 0.6611 & -- & -- & -- \\
        \midrule
        \textbf{Mean (matched five)} & \textbf{0.3601} & \textbf{0.4640} & -- & \textbf{0.6593} & \textbf{0.7181} & \textbf{0.7438} & \textbf{0.8328} & \textbf{0.7537} \\
        \bottomrule
    \end{tabular}
\end{table*}

The constant baselines show that the low raw scores are not uniformly substantive classification: MultiOFF is below its calibration-selected majority baseline, while HarMeme equals its constant-negative baseline. Threshold calibration nevertheless improves all six native margins. Across the five matched tasks, native calibration recovers
\[\frac{0.6593-0.4640}{0.7438-0.4640} = 0.698\]
of the raw native-to-probe gap. Routing then reaches \(0.7537\), adding \(0.0944\) over calibrated native scoring. It exceeds the probe alone on FHM and MAMI and is nearly equal on HarMeme, indicating complementarity rather than merely replacing a shifted threshold. The selected routing coefficients are 8 for FHM, CrisisHateMM, MAMI, and MMHS150K and 16 for HarMeme, all interior to the tested grid \(\{0,\ldots,64\}\).

\begin{table}[ht]
    \centering
    \caption{Recorded configuration of the dedicated and shared jointly supervised LoRA runs.}
    \label{tab:lora_run_configuration}
    \small
    \begin{tabular}{@{}
        >{\raggedright\arraybackslash}p{0.25\textwidth}
        >{\raggedright\arraybackslash}p{0.32\textwidth}
        >{\raggedright\arraybackslash}p{0.35\textwidth}
    @{}}
        \toprule
        Configuration & FHM-dedicated adapter & Shared seven-task adapter \\
        \midrule

        Base model & Gemma-3-4B-IT & Gemma-3-4B-IT \\
        Training tasks & FHM binary hatefulness & FHM, MAMI, MultiOFF, CrisisHateMM-A, CrisisHateMM-B, HarMeme, and MMHS150K \\

        Available target rows & \(7{,}938\) & \(160{,}761\) across seven tasks \\
        Optimization steps & \(2{,}000\) & \(4{,}000\) \\

        Task-sampling schedule & Single-task sampling & Equal round-robin task exposure; each task receives \(1/7\) of task-selection steps \\
        LoRA rank \(r_{\mathrm{LoRA}}\) & \(16\) & \(16\) \\
        Adapted modules & Seven projection types in all \(34\) transformer layers & Same seven projection types across all 34 layers \\

        Trainable parameters & \(29{,}802{,}496\) of \(4{,}329{,}881{,}968\) parameters (\(0.6883\%\)) & $29,802,496$ of $4,329,881,968$ parameters ($0.6883$\%) \\
        Gold-label loss & Cross-entropy; \(\lambda_{\mathrm{gold}}=0.5\) & Cross-entropy; \(\lambda_{\mathrm{gold}}=0.5\) \\

        Probe-target loss & KL distillation from the frozen FHM probe; \(\lambda_{\mathrm{probe}}=0.5\) & KL distillation from the corresponding frozen task probe; \(\lambda_{\mathrm{probe}}=0.5\) \\

        Pair-ranking loss & Active on \(4{,}330\) pseudo-confounder pairs; \(\lambda_{\mathrm{pair}}=0.5\) & Active only for tasks with valid pairs; \(\lambda_{\mathrm{pair}}=0.5\) \\

        Pair margin \(\delta\) & $1.0$ & $1.0$ \\
        Pairs sampled per eligible task per step & 4 FHM pairs & \(1\) pair, reduced from \(2\) after the initial run \\

        Pair-loss accumulation & Four sampled pair losses are averaged into the final microbatch loss, then scaled by $1/n_{acc}$ & Each sampled pair is backpropagated separately with effective weight \(\lambda_{\mathrm{pair}}/(n_{\mathrm{valid}}n_{\mathrm{acc}})\), where \(n_{\mathrm{valid}}\) is the number of valid sampled pairs and \(n_{\mathrm{acc}}\) is the gradient-accumulation factor \\

        Output-rate regularization & None & Task-dependent; see Table~\ref{tab:lora_task_sampling} \\
        Output-rate coefficient \(\lambda_{\mathrm{rate}}\) & \(0\) & $0.5$; EMA decay $0.98$ \\

        Batch size & $1$ & $1$ \\
        Gradient-accumulation factor \(n_{\mathrm{acc}}\) & $8$ & $8$ \\
        Effective main-loss batch per optimizer step & 8 examples & 8 examples \\
        Optimizer and learning rate & AdamW, $lr=10^{-4}$, weight decay $0$; $50$-step linear warmup & AdamW, $lr=10^{-4}$, weight decay $0$; $100$-step linear warmup \\

        \bottomrule
    \end{tabular}
\end{table}

\subsection{Jointly Supervised LoRA}
\label{app:probe-distillation}

Let \(p_{\theta,\psi,d}(\cdot\mid x)\) be the label distribution of the adapted LVLM, where \(\theta\) is frozen and \(\psi\) contains the LoRA parameters, and let \(q_d(\cdot\mid x)\) be the frozen probe
distribution. The task-\(d\) objective is
\begin{equation}
    \begin{aligned}
    \mathcal{L}^{(d)} = {}& \lambda_{\mathrm{gold}}\operatorname{CE}\left(y,p_{\theta,\psi,d}(\cdot\mid x)\right) + \lambda_{\mathrm{probe}}\operatorname{KL}\left(q_d(\cdot\mid x)\;\|\;p_{\theta,\psi,d}(\cdot\mid x)\right)
    \\
    &+ \mathbb{I}[|\mathcal{P}_d|>0]\,\lambda_{\mathrm{pair}}\mathcal{L}^{(d)}_{\mathrm{pair}}
    + \mathbb{I}[d\in\mathcal{D}_{\mathrm{rate}}]\,\lambda_{\mathrm{rate}} \mathcal{L}^{(d)}_{\mathrm{rate}} .
    \end{aligned}
    \label{eq:app-distillation-objective}
\end{equation}

The gold-label term limits blind imitation of probe errors, while the KL term transfers the probe's graded class preferences. Here \(\operatorname{CE}\) is cross-entropy against the gold label \(y\), \(\operatorname{KL}\) is the Kullback--Leibler divergence, \(\mathbb{I}[\cdot]\) is the indicator function, \(\mathcal{P}_d\) is the set of valid matched opposite-label training pairs for task \(d\), \(\mathcal{D}_{\mathrm{rate}}\) is the set of tasks with an output-rate regularizer \(\mathcal{L}^{(d)}_{\mathrm{rate}}\) (Table~\ref{tab:lora_task_sampling}), and the \(\lambda\) terms are loss weights (Table~\ref{tab:lora_run_configuration}). For tasks with matched opposite-label pairs,
\begin{equation}
    \mathcal{L}^{(d)}_{\mathrm{pair}} = \frac{1}{|\mathcal{P}_d|}\sum_{(x^+,x^-)\in\mathcal{P}_d}\left[
    \delta - m^{(d)}_{\theta,\psi}(x^+) + m^{(d)}_{\theta,\psi}(x^-)\right]_+
    \label{eq:app-distillation-pair-loss}
\end{equation}

where \(m^{(d)}_{\theta,\psi}\) is the adapted model's signed harmful-versus-benign label margin, \(x^+\) and \(x^-\) are the harmful and benign pair members, respectively, \(\delta\) is the ordering margin, and \([a]_+=\max(0,a)\).

\paragraph{Dedicated FHM adapter.}
The dedicated Gemma-3-4B-IT run uses rank-16 LoRA updates and contains $29,802,496$ trainable parameters, corresponding to \(0.6883\%\) of the 4.33-billion-parameter model. It is trained for $2,000$ steps using $7,938$ available target rows and $4,330$ pseudo-confounder pairs. On the same $1000$ scored test-seen examples, the base model, gold-only adapter, and joint gold-plus-KL-plus-pair adapter obtain \(0.6482\), \(0.6462\), and \(0.7138\) macro-F1, respectively. The joint objective therefore improves FHM beyond gold supervision alone without requiring an SAE or probe at inference.

\paragraph{Shared seven-task adapter.}
The shared adapter is trained for $4,000$ steps with equal round-robin task sampling, so each task receives one seventh of task-selection steps, irrespective of its share of the $160,761$ available target rows. Table~\ref{tab:lora_gold_control} compares it with a compute-matched gold-only adapter using the same target rows, rank, adapted modules, optimizer settings, steps, and sampling schedule. Both improve on the base mean, but gold-only reaches \(0.5780\), compared with \(0.5686\) for the joint objective. The proportional-sampling joint run reaches \(0.5374\), showing schedule sensitivity but not constituting a loss-matched ablation.

\begin{table}[ht]
    \centering
    \caption{\textbf{Gold-only LoRA control for the shared seven-task adapter.} Macro-F1 is measured with the same evaluator. Gold-only and Joint RR are compute-matched round-robin runs; Joint Prop. changes only the task-sampling schedule. Joint denotes gold cross-entropy plus probe KL, pair ranking where available, and task-specific output-rate regularization. A KL-only/no-pair condition was not run, so the individual auxiliary losses cannot be isolated.}
    \label{tab:lora_gold_control}
    \small
    \setlength{\tabcolsep}{4pt}
    \begin{tabular}{@{}lcccc@{}}
        \toprule
        Task & Base & Gold only & Joint RR & Joint Prop. \\
        \midrule
        FHM & 0.6567 & 0.6365 & 0.6548 & 0.6469 \\
        MAMI & 0.4859 & 0.8459 & 0.8620 & 0.8609 \\
        MultiOFF & 0.5187 & 0.6537 & 0.6373 & 0.5168 \\
        CrisisHateMM-A & 0.6058 & 0.8514 & 0.8182 & 0.7661 \\
        HarMeme (3-class) & 0.2889 & 0.4776 & 0.5078 & 0.4969 \\
        MMHS150K (6-class) & 0.3727 & 0.2688 & 0.1886 & 0.2019 \\
        CrisisHateMM-B & 0.3204 & 0.3122 & 0.3113 & 0.2723 \\
        \midrule
        \textbf{Mean} & \textbf{0.4642} & \textbf{0.5780} & \textbf{0.5686} & \textbf{0.5374} \\
        \bottomrule
    \end{tabular}
\end{table}

The auxiliary losses have task-dependent effects relative to gold-only: Joint RR improves FHM, MAMI, and HarMeme, but reduces MultiOFF, both CrisisHateMM tasks, and MMHS150K. The largest reduction is on MMHS150K, where six-class macro-F1 falls from \(0.2688\) under gold-only training to \(0.1886\) under the joint objective; both are below the \(0.3727\) base score. This is shared-adaptation interference rather than evidence that probe distillation uniformly transfers useful structure. These results favor task-specific or task-grouped adapters over a single universal adapter.




\begin{table}[ht]
    \centering
    \caption{Per-task target pools and auxiliary supervision for the shared adapter. Pool share is computed over the \(160{,}761\) available target rows. ``Configured, inactive'' means that the task was eligible for a pair loss but no valid training pairs were found.}
    \label{tab:lora_task_sampling}
    \small
    \setlength{\tabcolsep}{4pt}
    \begin{tabular}{@{}cccccc>{\raggedright\arraybackslash}p{0.23\textwidth}@{}}
        \toprule
        Task & Targets \(N_d\) & Pool share & Step share & Valid pairs & Pair loss & Output-rate regularizer \\
        \midrule

        FHM & \(7{,}938\) & \(4.938\%\) & \(1/7\) & \(4{,}330\) & Active & None \\

        MAMI & \(9{,}000\) & \(5.598\%\) & \(1/7\) & \(23\) & Active & Active; EMA prediction rate compared with the approximately \(0.5\) empirical positive-class prior \\

        MultiOFF & \(445\) & \(0.277\%\) & \(1/7\) & \(1\) & \shortstack[c]{Active, but\\ extremely\\ sparse} & Not used \\

        \shortstack[c]{CrisisHateMM\\-A} & \(3{,}600\) & \(2.239\%\) & \(1/7\) & \(0\) & \shortstack[c]{Configured,\\ inactive} & Not used \\

        HarMeme & \(3{,}013\) & \(1.874\%\) & \(1/7\) & \(51\) & Active & Active on the binary \texttt{harmful}/\texttt{not harmful} collapse; it does not regularize the \texttt{somewhat harmful}/\texttt{very harmful} boundary \\

        MMHS150K & \(134{,}823\) & \(83.865\%\) & \(1/7\) & \(57\) & Active & Active using a one-sided EMA overshoot statistic relative to the approximately \(0.583\) empirical Hate prior; the term was largely inactive because the EMA rate remained below the prior \\

        \shortstack[c]{CrisisHateMM\\-B} & \(1{,}942\) & \(1.208\%\) & \(1/7\) & \(0\) & Inactive & Not used because the three-way target task has no natural harmful-versus-benign output-rate constraint \\

        \midrule
        \textbf{Total} & \(\mathbf{160{,}761}\) & \(\mathbf{100.000\%}\) & -- & \(\mathbf{4{,}462}\) & -- & -- \\

        \bottomrule
    \end{tabular}
\end{table}




\section{Multilingual, OCR, and Visual-Credit Robustness Details}
\label{app:robustness-details}

\paragraph{Perturbation protocol.}
For prediction system \(S\in\{\mathrm{native},\mathrm{probe},\mathrm{route}\}\), each example
is evaluated under four conditions: \(\mathrm{orig}\) uses the original image and supplied OCR; \(\mathrm{blank}\) replaces the image with a uniform grey image; \(\mathrm{shuffle}\) assigns a deterministic unrelated image from the same reporting split; and \(\mathrm{noOCR}\) removes the supplied OCR string while retaining the original image. Let \(F^{(c)}_{S,d}\) be system \(S\)'s macro-F1 on dataset \(d\) under condition \(c\). We report
\begin{equation}
    \Delta^{\mathrm{OCR}}_{S,d} = F^{(\mathrm{orig})}_{S,d} - F^{(\mathrm{noOCR})}_{S,d},
    \qquad
    \Delta^{\mathrm{shuf}}_{S,d} = F^{(\mathrm{orig})}_{S,d} - F^{(\mathrm{shuffle})}_{S,d}.
    \label{eq:app-robustness-deltas}
\end{equation}

Aggregate score changes need not identify whether the affected examples were originally correct. For reporting example \(i\) with gold label \(y_i\), let \(\widehat{y}^{(c)}_{S,i}\) be system \(S\)'s prediction under condition \(c\), and let \(n_d\) be the number of reporting examples. We therefore define
\begin{equation}
    \begin{aligned}
    q^{\mathrm{img}}_{S,i} ={}& \mathbb{I}\left[\widehat{y}^{(\mathrm{orig})}_{S,i}=y_i\right]
    \mathbb{I}\left[\widehat{y}^{(\mathrm{blank})}_{S,i}\neq y_i\;\lor\; \widehat{y}^{(\mathrm{shuffle})}_{S,i}\neq y_i\right],
    \\
    C^{\mathrm{img}}_{S,d} ={}& \frac{\sum_{i=1}^{n_d}q^{\mathrm{img}}_{S,i}}{\sum_{i=1}^{n_d}\mathbb{I} \left[\widehat{y}^{(\mathrm{orig})}_{S,i}=y_i\right]}.
    \end{aligned}
    \label{eq:app-image-credit}
\end{equation}

Thus, with \(\mathbb{I}[\cdot]\) the indicator function and \(\lor\) logical disjunction, \(C^{\mathrm{img}}_{S,d}\) is the fraction of originally correct predictions that become incorrect under at least one image perturbation. It measures operational image dependence; it does not allocate causal credit between modalities.

\paragraph{Cross-lingual decodability on EXIST.}
We use the deterministic internal holdouts defined in Appendix~\ref{app:data-eval}~\citep{10.1007/978-3-032-04354-2_16}. With Qwen's public base SAE, an English-trained all-token probe reaches \(0.7469\) macro-F1 on Spanish, while a Spanish-trained generated-token probe reaches \(0.7739\) on English. Bilingual training reaches \(0.8121\) on English using prompt/OCR states and \(0.7623\) on Spanish using all-token states. The strongest role changes across transfer directions, but decodability stays high outside the probe's training language.

Gemma gives a less stable result. Across five independently constructed holdouts, every analyzed representation remains above random prediction, but each transfer direction contains at least one split on which the probe-minus-native difference is negative, and the standard deviation across splits exceeds the mean gain. For Gemma, EXIST thus shows cross-lingual internal decodability but no stable improvement over native prediction.

\begin{table}[ht]
  \centering
  \small
  \setlength{\tabcolsep}{6pt}
    \caption{\textbf{EXIST transfer stability over split seeds (Gemma).} Mean\,$\pm$\, sample SD of the paired probe-minus-native macro-F1 gain over deterministic holdout seeds 13, 29, 42, 73, and 101, not initialization seeds or confidence intervals. English-to-Spanish uses all-token states; the other two settings use generated-token states.}
  \label{tab:exist-transfer-seed-summary}
  \begin{tabular}{@{}lr@{}}
    \toprule
    Transfer direction & Gain (macro-F1) \\
    \midrule
    English$\rightarrow$Spanish & $0.0145\pm0.0302$ \\
    Spanish$\rightarrow$English & $0.0095\pm0.0373$ \\
    Bilingual (generated-token) & $0.0139\pm0.0213$ \\
    \bottomrule
  \end{tabular}
\end{table}

\paragraph{Code-mixed decodability on MultiBully.}
On Hindi-English code-mixed MultiBully \citep{10.1145/3477495.3531925}, Qwen's all-token base-SAE probe obtains \(0.7259\) macro-F1, compared with \(0.3401\) for native constrained output. Gemma's all-token residual-SAE probe obtains \(0.7109\) on the primary split (seed 42), compared with \(0.5990\) natively. Across five deterministic label-stratified 1,000-example holdouts (Table~\ref{tab:multibully-gemma-seeds}), its macro-F1 is \(0.6892\pm0.0143\); it outperforms native prediction on every split, and the mean paired difference is \(0.0915\pm0.0226\), with uncertainties given as sample standard deviations across split seeds. Shuffled-label controls remain near chance (\(0.4745\)--\(0.5429\) macro-F1).

\begin{table}[ht]
    \centering
    \small
    \setlength{\tabcolsep}{6pt}
    \caption{\textbf{MultiBully split stability (Gemma-3-4B).} Macro-F1 of the all-token residual-SAE probe and native constrained yes/no prediction on five deterministic label-stratified 1,000-example holdouts ($4{,}793$ training examples each). Seeds define holdout splits, not model initialization.}
    \label{tab:multibully-gemma-seeds}
    \begin{tabular}{@{}rccc@{}}
        \toprule
        Split seed & Probe & Native & \(\Delta\) \\
        \midrule
        13 & 0.6768 & 0.6176 & \(+0.0592\) \\
        29 & 0.6796 & 0.6008 & \(+0.0788\) \\
        42 & 0.7109 & 0.5990 & \(+0.1119\) \\
        73 & 0.6967 & 0.5849 & \(+0.1118\) \\
        101 & 0.6822 & 0.5865 & \(+0.0957\) \\
        \midrule
        Mean \(\pm\) SD & \(0.6892\pm0.0143\) & \(0.5978\pm0.0132\) & \(+0.0915\pm0.0226\) \\
        \bottomrule
    \end{tabular}
\end{table}

\paragraph{Multilingual causal replication.}
Using the silent-feature intervention protocol from Appendix~\ref{app:causal-details}, Qwen gives probe-to-native sensitivity ratios of approximately \(62{:}1\) on English EXIST, \(90{:}1\) on Spanish EXIST, and \(48.5{:}1\) on MultiBully. The two EXIST estimates use 12 probe-correct positive examples per language, while MultiBully uses 15 examples. These small intervention sets support a mechanistic
replication of the probe-dominant pattern.

\begin{table}[ht]
    \centering
    \caption{\textbf{Qwen visual-credit audit on disjoint reporting partitions.} \(P\) denotes the frozen public base-SAE probe and \(R\) the calibration-only logit router. \(\Delta^{\mathrm{shuf}} = F^{(\mathrm{orig})}-F^{(\mathrm{shuffle})}\). The final column reports the conditional image-credit rates \(C^{\mathrm{img}}_P/C^{\mathrm{img}}_R\). All probe and router values within a row use identical reporting examples; they differ from full-split headline results because router calibration uses a separate partition.}
    \label{tab:visual-credit-audit}
    \small
    \setlength{\tabcolsep}{3.3pt}
    \renewcommand{\arraystretch}{0.96}
    \begin{tabular}{@{}lrrrrrr@{}}
        \toprule
        Dataset & \(n_d\) & \(F^{\mathrm{orig}}_P\) & \(\Delta^{\mathrm{shuf}}_P\) & \(F^{\mathrm{orig}}_R\) & \(\Delta^{\mathrm{shuf}}_R\) & \(C^{\mathrm{img}}_P/C^{\mathrm{img}}_R\) \\
        \midrule
        FHM & 651 & 0.679 & 0.162 & 0.687 & 0.160 & 70.4/67.8\% \\
        MAMI & 700 & 0.664 & 0.206 & 0.801 & 0.292 & 88.8/64.7\% \\
        MultiBully & 700 & 0.618 & 0.148 & 0.720 & 0.200 & 65.8/69.1\% \\
        EXIST-ES & 175 & 0.729 & 0.182 & 0.713 & 0.178 & 55.8/64.0\% \\
        HarMeme & 248 & 0.698 & 0.162 & 0.782 & 0.211 & 56.7/62.6\% \\
        \bottomrule
    \end{tabular}
\end{table}

\paragraph{Supplied-OCR ablations.}
On Qwen EXIST, removing supplied OCR lowers prompt-probe macro-F1 by \(0.0507\) for English-to-Spanish transfer, \(0.0671\) for bilingual training evaluated on English, and \(0.0218\) for bilingual training
evaluated on Spanish. Spanish-to-English transfer is the exception: macro-F1 changes from \(0.7583\) to \(0.7684\) after OCR removal.

On MultiBully, the prompt-position probe loses \(0.0523\), the all-token probe loses \(0.0236\), and the image-position probe changes by only \(0.0011\). Supplied OCR helps text-bearing representations the most, but its effect is uneven and too small to explain the full signal. The no-OCR condition removes only the OCR string inserted into the prompt. Meme text remains visible in the image, and its removal can indirectly alter subsequent generated-token states, so this condition is not a text-free input.

\begin{table}[ht]
    \centering
    \small
    \caption{\textbf{Gemma-3-4B visual-credit classification controls.} Macro-F1 on fixed audit splits. FHM uses its dedicated LoRA; MAMI and HarMeme use the shared task-trained adapter; MultiBully and EXIST use that shared adapter zero-shot. All systems and conditions within a task use the same example count.}
    \label{tab:gemma-visual-credit}
    \begin{tabular}{@{}ccccccc@{}}
        \toprule
        Task / split & System / adapter & Target Count & Original & Blank & Shuffled & No-OCR \\
        \midrule
        \shortstack[c]{FHM /\\ validation} & Native & 469 & 0.657 & 0.585 & 0.553 & 0.650 \\
        & LoRA, dedicated & 469 & 0.710 & 0.502 & 0.517 & 0.696 \\
        & Probe & 469 & 0.701 & 0.591 & 0.476 & 0.656 \\
        \addlinespace
        \shortstack[c]{MAMI /\\ test} & Native & 1,000 & 0.409 & 0.354 & 0.360 & 0.416 \\
        & LoRA, shared in-domain & 1,000 & 0.818 & 0.700 & 0.537 & 0.812 \\
        & Probe & 1,000 & 0.694 & 0.333 & 0.482 & 0.703 \\
        \addlinespace
        \shortstack[c]{HarMeme /\\ validation} & Native & 177 & 0.462 & 0.475 & 0.440 & 0.492 \\
        & LoRA, shared in-domain & 177 & 0.792 & 0.715 & 0.553 & 0.775 \\
        & Probe & 177 & 0.710 & 0.396 & 0.487 & 0.727 \\
        \addlinespace
        \shortstack[c]{MultiBully /\\ internal} & Native & 1,000 & 0.599 & 0.594 & 0.549 & 0.538 \\
        & LoRA, shared zero-shot & 1,000 & 0.547 & 0.448 & 0.445 & 0.523 \\
        & Probe & 1,000 & 0.711 & 0.323 & 0.539 & 0.681 \\
        \addlinespace
        \shortstack[c]{EXIST-ES /\\ internal} & Native & 500 & 0.658 & 0.608 & 0.547 & 0.660 \\
        & LoRA, shared zero-shot & 500 & 0.648 & 0.581 & 0.511 & 0.591 \\
        & Probe & 500 & 0.600 & 0.272 & 0.490 & 0.596 \\
        \bottomrule
    \end{tabular}
\end{table}

\paragraph{Visual-credit results.}
As shown in Table~\ref{tab:visual-credit-audit}, image permutation reduces probe macro-F1 by \(0.148\)--\(0.206\) and routed macro-F1 by \(0.160\)--\(0.292\) on every audited dataset. Conditional image credit ranges from \(55.8\%\) to \(88.8\%\) for the probe and from \(62.6\%\) to \(69.1\%\) for the router. The corresponding native-model rate is only \(0.0\%\)--\(38.1\%\), partly because several native classifiers have collapsed or strongly biased output distributions. Post-hoc classification of supplied OCR cannot account for the probe advantage.

\paragraph{Gemma visual-credit and prompt-prior controls.}
Table~\ref{tab:gemma-visual-credit} complements the Qwen probe-router audit with native, probe, and LoRA results for Gemma. Image shuffling produces substantially larger losses than OCR removal for the strongest systems, while the blank-image results show that visual credit is material but not exclusive. The shared zero-shot adapter does not outperform native Gemma on MultiBully ($0.547$ versus $0.599$) or EXIST-ES ($0.648$ versus $0.658$); neither is a task-specific LoRA result. The altered-prompt conditions in Table~\ref{tab:gemma-prompt-prior} are reported only as prompt-prior diagnostics.

\begin{table}[ht]
    \centering
    \small
    \caption{\textbf{Gemma prompt-prior diagnostics, not classification baselines.} Each cell lists native / LoRA / probe macro-F1 under a changed prompt. Splits, sample counts, and adapters match Table~\ref{tab:gemma-visual-credit}. These prompts do not preserve the classification instruction, so their scores cannot isolate OCR or visual contributions to task classification.}
    \label{tab:gemma-prompt-prior}
    \begin{tabular}{@{}lcc@{}}
        \toprule
        Task & Blank + OCR-only prompt & Neutral-description prompt \\
        \midrule
        FHM & 0.406 / 0.395 / 0.338 & 0.338 / 0.585 / 0.558 \\
        MAMI & 0.379 / 0.433 / 0.333 & 0.333 / 0.433 / 0.699 \\
        HarMeme & 0.285 / 0.623 / 0.396 & 0.349 / 0.396 / 0.715 \\
        MultiBully & 0.422 / 0.438 / 0.480 & 0.355 / 0.447 / 0.600 \\
        EXIST-ES & 0.473 / 0.441 / 0.272 & 0.385 / 0.387 / 0.594 \\
        \bottomrule
    \end{tabular}
\end{table}

\paragraph{FHM matched-pair controls.}
On opposite-label FHM pairs sharing the same image, Qwen's probe and router assign the hateful member the larger score in \(86.4\%\) and \(88.4\%\) of cases, respectively. On pairs sharing the same text but
using different images, the corresponding rates are \(83.8\%\) and \(86.3\%\). These ordering results show sensitivity to the component that changes within each pair, although correctly classifying both
members remains more difficult.

Gemma independently reproduces the qualitative visual-sensitivity pattern. Image shuffling reduces macro-F1 by approximately \(0.17\)--\(0.28\) for its strongest FHM, MAMI, HarMeme, and MultiBully
probe or LoRA systems.

\paragraph{Blanking versus shuffling.}
Blank and shuffled images are distinct out-of-distribution interventions. Blanking removes most natural visual content and can change a model's class prior, whereas shuffling supplies a plausible but semantically mismatched image. This distinction explains cases such as the Qwen HarMeme probe, which improves slightly under blanking but degrades substantially under shuffling. We therefore use the two
conditions jointly to establish visual dependence and do not interpret either as identifying a specific cross-modal circuit~\citep{hessel-lee-2020-multimodal}.

\section{Qualitative Feature Examples}
\label{app:feature-examples}

\paragraph{Selection and interpretation.}
Before rendering the revised atlas, we fix a retrospective four-feature case-study set: $02261$ as an FHM group-antagonism feature included in the causal audit, $01867$ as a CrisisHateMM domain contrast, $02205$ as a MAMI relationship contrast, and $02629$ as a HarMeme feature also listed in the FHM confounder audit. This is not a preregistered or globally highest-$|w|$ selection. We report actual fitted weights and their absolute ranks instead of claiming that these four are the most probe-important features.


For each feature, we select the two highest max-pooled activations within its source task's fixed validation pool, with sample ID breaking ties and no filtering by label or image appearance. Exemplars and probe coefficients use the same token role: all tokens for the saved FHM probe and image tokens for the other three probes. Signed weights refer to the stated class on the probe's scaled inputs; ranks are among all $20,480$ coefficients of that class. Confounder ranks and causal inclusion come from separate, explicitly role-labeled audits, not from these validation exemplars.

The matched examples narrow the qualitative claims: $01867$ illustrates conflict-related context, but $02205$'s strongest displayed MAMI examples are benign, and its coefficient for misogyny is negative. Likewise, $02629$'s displayed HarMeme examples are non-harmful, and its coefficient for very harmful is small. These cards show contextual features; none of them is a pure harmfulness detector. Only $02261$ has documented inclusion in the saved causal-card audit at image positions; that does not establish causal efficacy for its all-token probe shown here. 


\begin{figure}[ht]
  \centering
    \includegraphics[width=\linewidth]{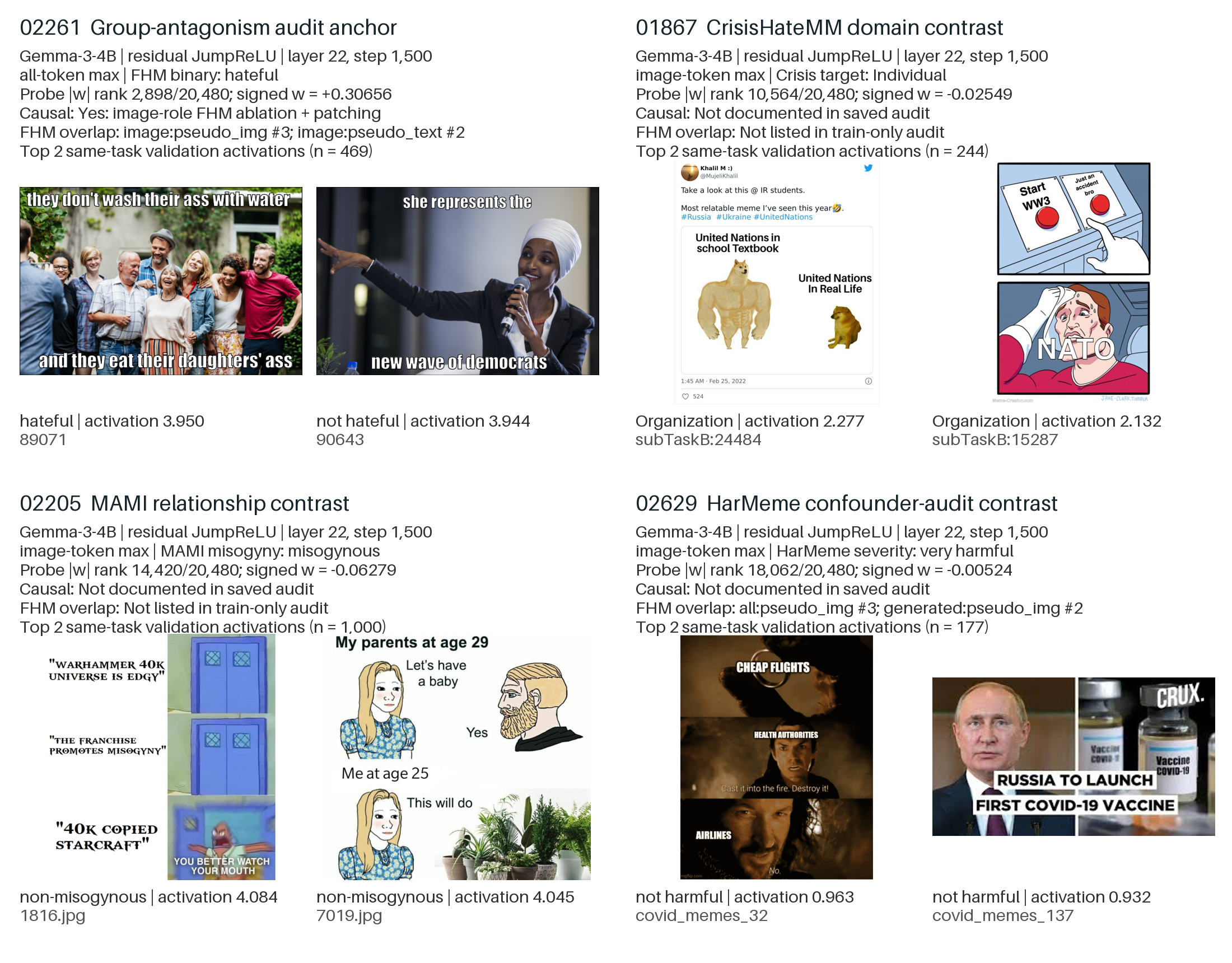}
    \caption{\textbf{Four-feature qualitative atlas.} Same-dataset, role-matched validation maxima for the selected Gemma-3-4B layer-22 residual SAE. Each card reports its source task and class, token role, fitted probe rank and signed weight, causal-audit inclusion, and train-only confounder-overlap status. The class naming the coefficient need not be the gold class of a top-activating example. Feature descriptions are qualitative context summaries, not claims of monosemanticity or complete explanations of model behavior.}
    \label{fig:feature-card-atlas}
\end{figure}

\begin{figure}[ht]
    \centering
    \includegraphics[width=0.9\linewidth]{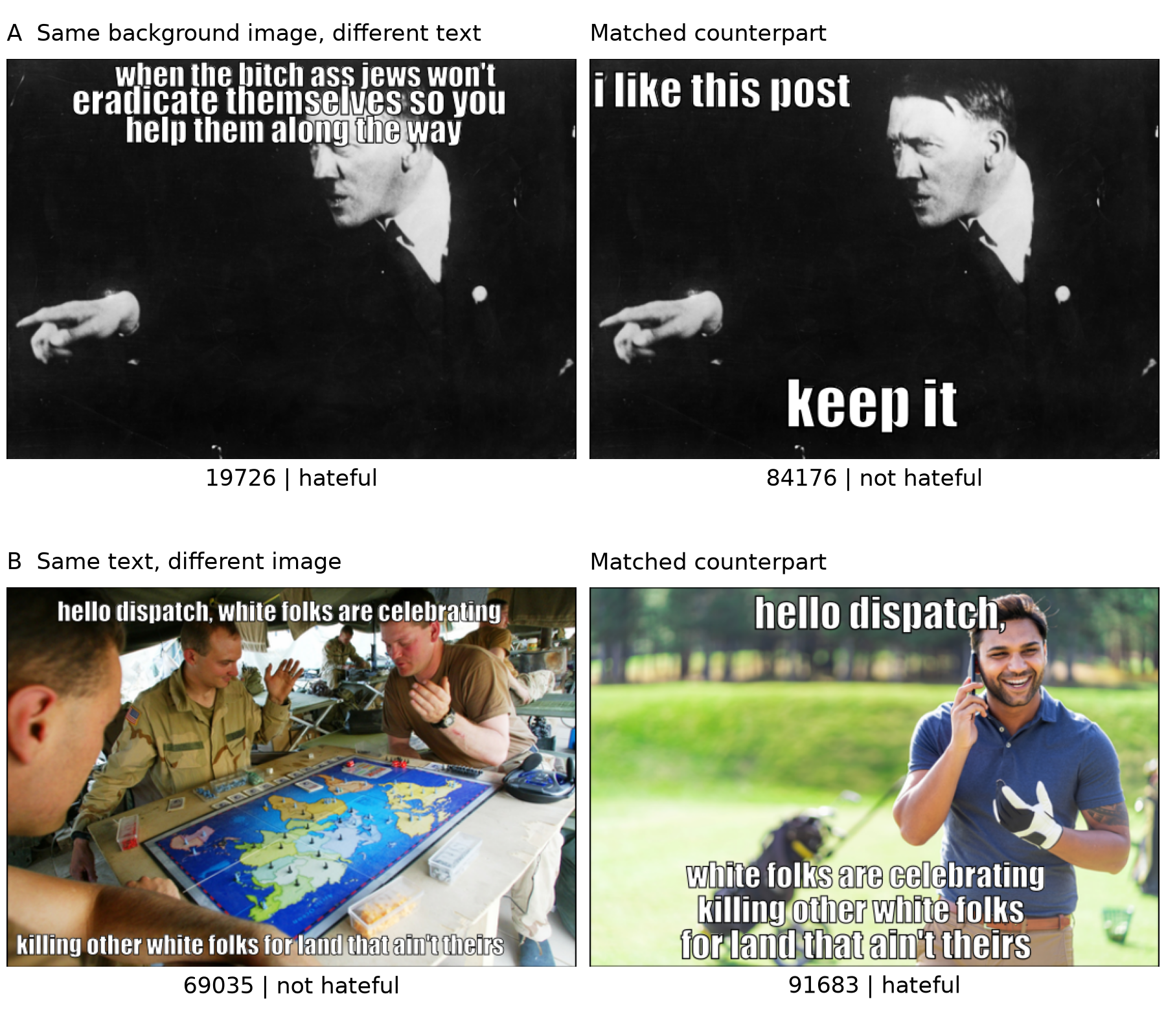}
    \caption{\textbf{Original FHM image and text confounders associated with feature 02261.} \textbf{A.} Training examples $19726$ and $84176$ share a background image but differ in text and gold label (same-image group 867). \textbf{B.} Examples $69035$ and $91683$ share the supplied text but differ in image and gold label (same-text group 952). These illustrative pairs are not selected from the validation maxima in the atlas and do not establish that feature $02261$ alone distinguishes either pair.}
    \label{fig:fhm-original-pairs}
\end{figure}

\paragraph{Original FHM confounder examples.}
Figure~\ref{fig:fhm-original-pairs} shows two matched pairs from the original $02261$ card. The anchors were retrieved among that feature's generated-token training exemplars, whereas the revised atlas reports all-token FHM validation activations. We keep these sources separate: the pairs illustrate the dataset's compositional contrasts, not new role-matched activation measurements or causal effects. The labels are the dataset annotations.

\section{Glossary of Terms}
\label{app:glossary}

This glossary defines terms according to their operational meaning in this paper. In particular, \emph{internal information} means information that a specified readout can recover on held-out data; it does not imply human-like knowledge, belief, or native model use.

\subsection{Representations and token roles.}

\glsentry{Native LVLM / native pathway}
{The original Gemma or Qwen model and its own route from the input,
through the transformer and output head, to the final label. No external
probe, activation hook, output router, or trained adapter determines the
prediction.}

\glsentry{Residual stream}
{The sequence of hidden-state vectors passed through the transformer
layers. We denote the activation at layer \(\ell\) and token position
\(t\) by \(h_{\ell,t}\). It should not be confused with an SAE
reconstruction residual.}

\glsentry{Sparse autoencoder (SAE)}
{An encoder--decoder model that maps a dense activation \(h\) to a sparse
latent vector \(z=E(h)\) and reconstructs it as
\(\widehat{h}=D(z)\)~\citep{ICLR2024_1fa1ab11,ICLR2025_42ef3308}.}

\glsentry{Sparse feature or SAE latent}
{One coordinate \(z_j\) in the SAE representation. Each feature has an
activation value and a decoder direction \(d_j\) in the LVLM's hidden
space. A sparse feature is not necessarily a perfectly monosemantic
human-interpretable concept.}

\glsentry{Feature dictionary}
{The collection of SAE decoder directions
\(\{d_1,\ldots,d_m\}\). Feature indices are specific to a particular
model, layer, and SAE; the same index in two dictionaries need not
represent the same pattern.}

\glsentry{Base SAE}
{The public SAE applied directly to an LVLM hidden state. The term
\emph{base} refers to its position in our SAE pipeline, not to the base
LVLM itself.}

\glsentry{Reconstruction residual}
{The information left after the base SAE reconstructs a hidden state:
\[
r_{\ell,t}
=
h_{\ell,t}-\widehat{h}^{\,b}_{\ell,t}.
\]
This is an SAE reconstruction error, not the transformer's residual
stream.}

\glsentry{Residual SAE}
{A second SAE trained on the reconstruction residual \(r_{\ell,t}\). It
captures structure not reconstructed by the base dictionary and is
treated as an alternative representation rather than an automatically
superior one.}

\glsentry{Joint reconstruction}
{The sum of the base-SAE reconstruction and decoded residual-SAE
reconstruction:
\[
\widehat{h}^{\,\mathrm{joint}}
=
\widehat{h}^{\,b}+\widehat{r}.
\]
This is the activation used by the residual reconstruction hook.}

\glsentry{Crosscoder}
{A sparse model that constructs one latent representation from several
transformer layers and uses separate decoders for those layers. It is
intended to organize information shared across model depth rather than
simply concatenate independent layer representations
\citep{lindsey2024crosscoders}.}

\glsentry{Token role}
{The provenance of a sequence position. We distinguish
\emph{prompt/OCR}, \emph{image-origin}, and \emph{generated} positions.
Token role identifies where a state originated, not which modality it
contains after contextual processing.}

\glsentry{Prompt/OCR tokens}
{Positions containing the task instruction, class wording, and supplied
OCR. At intermediate layers, these states may already contain visually
conditioned information.}

\glsentry{Image-position states}
{Positions originating from the visual encoder. Because they interact
with prompt states inside the transformer, they should not be interpreted
as modality-pure or strictly image-only representations.}

\glsentry{Generated-token states}
{Hidden states produced after answer generation begins. They may encode
the model's emerging label, explanation, or confidence. Gold labels are not supplied during their extraction, but the states may encode the model's predicted answer and are not pre-decision representations.}

\glsentry{Pre-generation representation}
{A representation pooled over prompt and image positions while excluding
generated positions. It tests what can be decoded before the model starts
producing its answer.}

\glsentry{All-token representation}
{A representation pooled over prompt, image, and generated positions. It
provides a full-computation decodability estimate, but is not a purely
pre-decision representation.}

\glsentry{Pooling}
{The operation that converts token-level activations into one vector per
meme. Our default is featurewise max pooling,
\[
[\phi_s(x)]_j
=
\max_{t\in\mathcal{T}_s(x)}z_{t,j},
\]
where \(s\) denotes the selected token role. Pooling discards token order
and exact activation position.}

\subsection{Readouts and output pathways.}

\glsentry{Readout}
{Any mapping from a representation to a task score or prediction. The
LVLM output pathway is a native readout; a probe is an external readout;
and the output router combines the two.}

\glsentry{Native prediction}
{The label obtained from the frozen LVLM using the task's constrained
decoding or label-scoring rule. \emph{Native} does not necessarily mean
free-form generation.}

\glsentry{Output head / unembedding}
{The final transformation from the model's hidden state to vocabulary
logits. Its directions indicate how changes in hidden space immediately
affect candidate output tokens.}

\glsentry{Native decision margin}
{A signed difference between competing native scores. For a literal
yes/no task,
\[
m_{\mathrm{yn}}(x)
=
\ell_{\mathrm{yes}}(x)-\ell_{\mathrm{no}}(x),
\]
where \(\ell_{\mathrm{yes}}\) and \(\ell_{\mathrm{no}}\) are the two
output logits. A positive value favors the designated positive class.}

\glsentry{External probe}
{A supervised classifier trained on frozen internal representations. A
probe measures how accessible a label is to the selected classifier
family; it does not establish that the native model uses the same
decision rule~\citep{hewitt2019designing,belinkov-2022-probing}.}

\glsentry{Sparse readout / SAE probe}
{A probe whose input is a pooled vector of SAE feature activations. The
input is sparse-derived, although the fitted classifier weights need not
themselves be sparse.}

\glsentry{Decodability}
{The extent to which a specified readout can recover a target from a
frozen representation on held-out data. Decodability is relative to the
chosen layer, token role, representation, probe family, and training
set.}

\glsentry{Internal signal}
{Shorthand for label-associated structure that is decodable from a model
representation. It does not imply that the model's native output pathway
uses that structure.}

\glsentry{Probe ceiling}
{The best held-out performance obtained by the selected probe and
representation. It is an empirical reference point, not a theoretical
upper bound on all possible classifiers.}

\glsentry{Internal-to-output gap / readout gap}
{The difference between probe and native performance on the same
reporting examples:
\[
\Delta_{\mathrm{readout}}
=
F_{\mathrm{probe}}-F_{\mathrm{native}}.
\]
Because the probe receives supervised labels while the native LVLM
remains frozen, this measures supervised accessibility rather than a
like-for-like model comparison.}

\glsentry{Hook}
{A mechanism attached to an intermediate layer to record or modify an
activation during the model's forward pass. The term identifies where an
intervention occurs, not the type of classifier used afterward.}

\glsentry{Residual reconstruction hook}
{An intervention that moves the original hidden state toward its joint
SAE reconstruction:
\[
h'_{\ell,t}
=
h_{\ell,t}
+
\alpha_{\mathrm{hook}}
\left(
\widehat{h}^{\,\mathrm{joint}}_{\ell,t}
-h_{\ell,t}
\right).
\]
The modified state then passes through the remaining LVLM layers and its
native output head.}

\glsentry{Direct logit routing / output router}
{An output-level combination of the native margin and standardized probe
score:
\[
m_d^{\mathrm{route}}(x)
=
m_d(x)+\beta_d\widetilde{u}_d(x),
\]
where \(m_d(x)\) is the native task margin,
\(\widetilde{u}_d(x)\) is the calibrated probe score, and \(\beta_d\) is
selected on a calibration set. No hidden state is modified.}

\glsentry{Probe distillation}
{Training the LVLM to reproduce useful probe behavior using gold labels
and soft probe targets. Unlike direct routing, the resulting model no
longer requires the SAE or external probe at inference.}

\glsentry{LoRA adapter}
{A parameter-efficient update that freezes an original weight matrix
\(W_0\) and learns a low-rank correction,
\[
W
=
W_0+\frac{\gamma}{r_{\mathrm{LoRA}}}BA,
\]
where \(A\) and \(B\) are trainable matrices,
\(r_{\mathrm{LoRA}}\) is the adapter rank, and \(\gamma\) is a scaling
factor~\citep{hu2021lora}.}

\subsection{Discriminative and routed directions.}

\glsentry{Discriminative direction}
{A feature or direction that helps the supervised probe separate task
labels. Discriminative importance does not imply influence on the
native output.}

\glsentry{Routed direction}
{A feature direction selected because it has a large projection onto the
native output direction. It may strongly affect the output margin without
being predictive of the correct label.}

\glsentry{Feature-to-output alignment}
{The direct projection
\[
a_j=d_j^\top u_{\mathrm{out}},
\]
where \(d_j\) is the decoder direction of feature \(j\) and
\(u_{\mathrm{out}}\) is the relevant native output direction. It is a
static approximation to immediate logit influence, not a complete causal
effect.}

\glsentry{Silent feature}
{A probe-important feature whose direct output alignment
\(|a_j|\) lies below a fixed threshold. \emph{Silent} means weakly
connected to the particular output channel being tested, not globally
inactive or causally irrelevant.}

\glsentry{Readout or routing misalignment}
{A situation in which the directions most useful to an external probe
differ from those that control the native output. This provides one
possible explanation for high decodability but weak native performance.}

\glsentry{Output anchor}
{The output contrast used to define a routed direction, such as
\texttt{yes} minus \texttt{no}. An anchor is mismatched when the task
actually uses a different rule, such as full-label string scoring.}

\glsentry{Donor and target}
{The \emph{donor} is the example from which an activation is taken.
The \emph{target} is the example whose computation is modified by that
activation.}

\glsentry{Ablation / knockout}
{An intervention that removes selected components, usually by setting
their feature activations to zero. Probe-side ablation tests dependence
of the external classifier; online ablation tests effects on the LVLM's
subsequent computation.}

\glsentry{Activation patching}
{Replacing part of a target example's internal representation with a
representation obtained from a donor example. Patching tests whether
transplanting a candidate feature pattern changes a downstream quantity.}

\glsentry{Sensitivity ratio}
{The ratio between an intervention's effect on the probe and its effect
on the native margin, or vice versa. It measures relative responsiveness under the chosen score scales, not a scale-invariant mediation fraction or whether the intervention improves correctness.}

\subsection{Cross-modal and robustness analyses.}

\glsentry{Benign confounder}
{An image--text combination that resembles a harmful example in one
modality but is benign when both modalities are interpreted together.
FHM was designed around such confounders~\citep{kiela2020hateful}.}

\glsentry{Matched confounder pair}
{Two memes that share one component, differ in the other, and have
opposite labels. A \emph{pseudo-image pair} shares an image but changes
the text; a \emph{pseudo-text pair} shares text but changes the image.}

\glsentry{Additive readout}
{A classifier of the form $f(x_{\mathrm{img}})+g(x_{\mathrm{prompt}})$, with no input-dependent cross-role coupling. Using both roles additively does not by itself demonstrate an interaction. Our gated pairwise baseline is not strictly additive because its gate depends on both role representations.}

\glsentry{Low-rank bilinear readout}
{A readout containing an explicit image-by-prompt term,
\[x_{\mathrm{img}}^\top UV^\top x_{\mathrm{prompt}}\]
where \(x_{\mathrm{img}}\) and \(x_{\mathrm{prompt}}\) are image and
prompt-position representations and \(U,V\in\mathbb{R}^{K\times
r}\). The rank \(r\) limits the number of
multiplicative interaction factors~\citep{kim2016hadamard}.}

\glsentry{Distributed interaction}
{An interaction whose predictive contribution is spread across several
bilinear factors. Small single-factor ablation effects show that no one tested factor is necessary for the full performance gain; they do not establish that every factor is individually insufficient or define a unique semantic factorization.}

\glsentry{Pair margin}
{The score difference
\[s(x^+)-s(x^-)\]
between harmful and benign members of a matched pair. A positive margin
means that the pair is ordered correctly even when one member crosses the
wrong hard-decision threshold.}

\glsentry{Positive-margin, separated, and both-correct rates}
{The positive-margin rate measures correct score ordering. The separated
rate measures whether the pair receives different hard labels. The
both-correct rate requires both members to receive their correct labels
and is therefore the strictest measure.}

\glsentry{No-OCR control}
{A condition in which supplied OCR is removed from the prompt while the
original image remains. Since meme text is still visible in the image,
this tests reliance on explicit OCR injection rather than providing a
text-free input.}

\glsentry{Visual credit}
{A paired measure of image dependence. An originally correct prediction
is image-credited when it becomes incorrect after image blanking or
same-split image shuffling. It is an operational robustness measure, not
a complete causal allocation of credit between modalities~\citep{liu2026visualcreditauditmultimodal}.}

\glsentry{Locked evaluation}
{A protocol in which the representation, token role, classifier,
threshold, router coefficient, and checkpoint are frozen before the
reporting split is scored. No reported examples are used for
post-hoc model selection.}

\end{document}